\documentclass[11pt]{article}

\usepackage[preprint]{acl}
\usepackage{multirow}
\usepackage{xcolor}
\usepackage{amsmath}
\usepackage{amssymb}
\usepackage{booktabs}
\usepackage{graphicx}
\usepackage[table]{xcolor}
\usepackage{arydshln}
\usepackage{subcaption}

\definecolor{PastelBlue}{rgb}{0.94, 0.97, 1.0}
\definecolor{Oursgray}{gray}{0.94}
\definecolor{DeltaGreen}{RGB}{34,139,34} 

\usepackage{times}
\usepackage{latexsym}

\usepackage[T1]{fontenc}

\usepackage[utf8]{inputenc}

\usepackage{microtype}

\usepackage{inconsolata}

\usepackage{graphicx}
\usepackage{subcaption}
\usepackage{amssymb}
\usepackage{amsmath}

\title{Is Discrete Difficulty Sufficient? Leveraging Continuous Difficulty for Efficient Self-Consistency in LLMs}

\author{
  \textbf{Sihyeong Yeom\textsuperscript{1,$\dagger$}},
  \textbf{Geon Park\textsuperscript{1,$\dagger$}},
  \textbf{Geunyeong Jeong\textsuperscript{1}},
  \textbf{Taewoong Yoon\textsuperscript{1}},
  \\
  \textbf{Jaewook Lee\textsuperscript{2}},
  \textbf{Harksoo Kim\textsuperscript{1,*}}
  \\
  \\
  \textsuperscript{1}Konkuk University
  \qquad
  \textsuperscript{2}DATUMO INC.
  \\
  \texttt{\{stv10121, albert0811, jyjg7218, twyoon816\}@konkuk.ac.kr}
  \\
  \texttt{benecia428@gmail.com, nlpdrkim@konkuk.ac.kr}
}

\begin{document}
\maketitle

\begingroup
\renewcommand{\thefootnote}{$\dagger$}
\footnotetext{Equal contribution.}
\endgroup

\begingroup
\renewcommand{\thefootnote}{*}
\footnotetext{Corresponding author.}
\endgroup

\begin{abstract}
Self-Consistency (SC) is a decoding strategy that samples diverse reasoning paths and selects the most consistent answer, demonstrating strong performance on complex reasoning problems. However, the excessive token consumption incurred by generating multiple reasoning paths has been identified as a major limitation of SC. To improve computational efficiency, several studies have proposed strategies that adjust the number of reasoning paths or allocate resources differentially according to problem difficulty. Nevertheless, most existing methods categorize difficulty into a few fixed levels, failing to fully capture the continuously varying nature of reasoning complexity. In this work, we propose \textbf{F}lexible \textbf{S}elf-\textbf{C}onsistency (\textbf{FSC}), which estimates problem difficulty as a continuous signal and dynamically adjusts the number of generated reasoning paths accordingly. FSC predicts the output entropy of an input question using a pre-trained probe and leverages it as an indicator of model uncertainty to flexibly control the sampling budget. Experimental results show that, across various models and benchmarks, FSC maintains accuracy comparable to SC while achieving token savings of up to 76\%.
\end{abstract}

\section{Introduction}
\label{sec: Intro}
\textbf{``\emph{Do all questions require the same number of reasoning paths?}''}---
Some questions converge to the correct answer reliably with only one or two reasoning attempts, whereas others reach the right conclusion only after exploring multiple possibilities. This simple fact that each problem requires a different amount of reasoning is creating an important turning point in recent language model research. Scaling train-time compute, such as model parameter size and training data size, has established Large Language Models (LLMs) as a core technology across natural language processing~\citep{scaling_laws, chinchilla, llm_survey, geunyeong, byungkook}. However, performance gains gradually diminish even when training-time compute continues to increase. In response, Test-Time Scaling (TTS), which allocates additional computation at inference time, has emerged as a promising alternative \citep{tts_survey, tts_2024, tts_2025}.

\begin{figure}[t]
\includegraphics[width=\columnwidth]{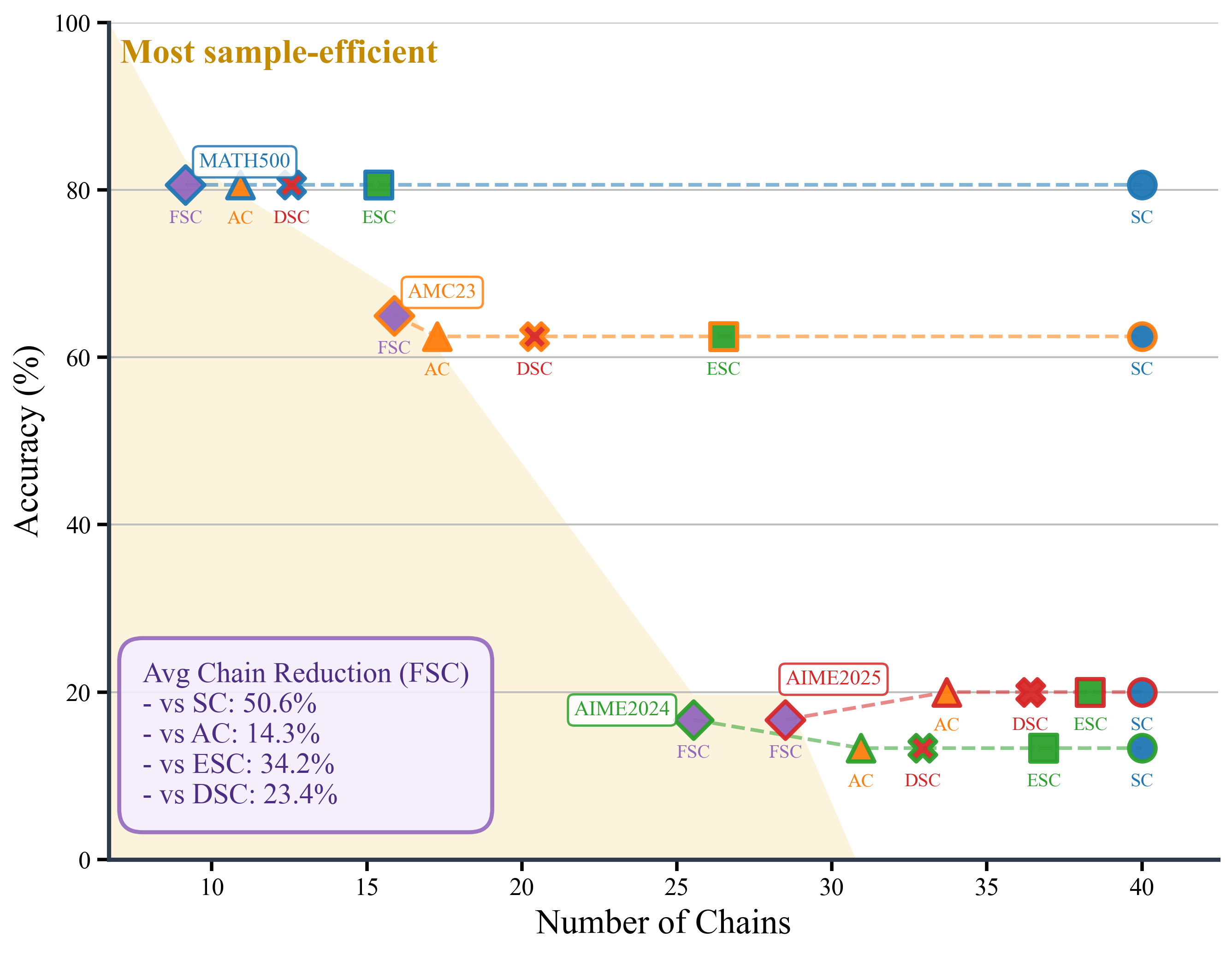}
\caption{\textbf{FSC achieves high efficiency with fewer reasoning chains.} Across datasets, comparison results between accuracy and the number of reasoning chains show that FSC tends to maintain comparable accuracy with fewer chains.}\label{fig:intro}
\end{figure}

TTS is an approach that improves model performance by performing additional computation during inference. Among such methods, Self-Consistency (SC) is a representative decoding strategy that samples multiple reasoning paths and aggregates them to determine the final answer~\citep{sc}. SC shows strong performance on tasks requiring complex reasoning, but as the number of generated paths increases, token usage also increases accordingly. In other words, although SC is effective for improving performance, how much reasoning resource should be allocated to obtain such gains remains a separate problem.

Prior work on improving the efficiency of SC has progressed from early stopping strategies to difficulty-adaptive reasoning allocation. Earlier studies improved token efficiency by checking the consistency or stability of responses during the reasoning process and stopping sampling early~\citep{ac, esc}. Subsequent work has advanced toward estimating the difficulty of an input question and dynamically adjusting the number of reasoning paths according to that difficulty~\citep{dsc, llm_already_knows}. In this way, the recent research trend provides a more direct solution in that it aims to assign computational resources according to the needs of each problem.

The key issue lies in how difficulty is represented. Existing difficulty-based methods generally classify difficulty into discrete categories such as `easy' and `hard' and assign a pre-defined number of reasoning paths to each category. However, even among problems belonging to the same category, there exist substantial differences in the amount of reasoning required, and these fine-grained differences are precisely what determine the balance between accuracy and computational efficiency. When difficulty is treated only coarsely, excessive computation may be allocated to some problems, while insufficient exploration may be provided to others. For example, assigning only a single path to easy problems and uniformly assigning 40 paths to hard problems fails to capture fine-grained difficulty differences among problems, ultimately making it difficult to precisely control reasoning resources.

At this point, we pose the following question:
\textbf{``\emph{Can problem difficulty be modeled not as a discrete category, but as a continuous signal?}''}

As an answer to this question, we propose \textbf{F}lexible \textbf{S}elf-\textbf{C}onsistency (\textbf{FSC}). Through preliminary experiments, we show that output entropy reflects the uncertainty exhibited by a model for a given input and can be used as a continuous difficulty signal. Based on this finding, FSC trains a lightweight linear probe to predict the output entropy of each input question and uses it as a continuous indicator of the amount of reasoning required by that question. Using this predicted value, FSC flexibly determines the number of reasoning paths needed for each question, allocating less computation to easy problems and more computation to difficult ones.

Experimental results on various benchmarks show that FSC substantially reduces token usage while maintaining accuracy comparable to SC. These results suggest that the proposed method is an efficient and general-purpose TTS strategy capable of more precisely controlling the balance between accuracy and efficiency.

In summary, this work makes the following key contributions.
\textbf{First}, we propose FSC, a new framework that adaptively allocates the number of reasoning paths required for each input based on predicted output entropy.
\textbf{Second}, we show that output entropy can serve as a continuous difficulty signal that replaces conventional discrete difficulty levels, and empirically demonstrate that entropy gradually increases with problem difficulty.
\textbf{Third}, we show that FSC can substantially improve inference efficiency over existing baselines while maintaining accuracy across diverse benchmarks.

\begin{figure*}[!t]
    \centering
    \begin{subfigure}[b]{0.59\linewidth}
        \centering
        \includegraphics[width=\linewidth]{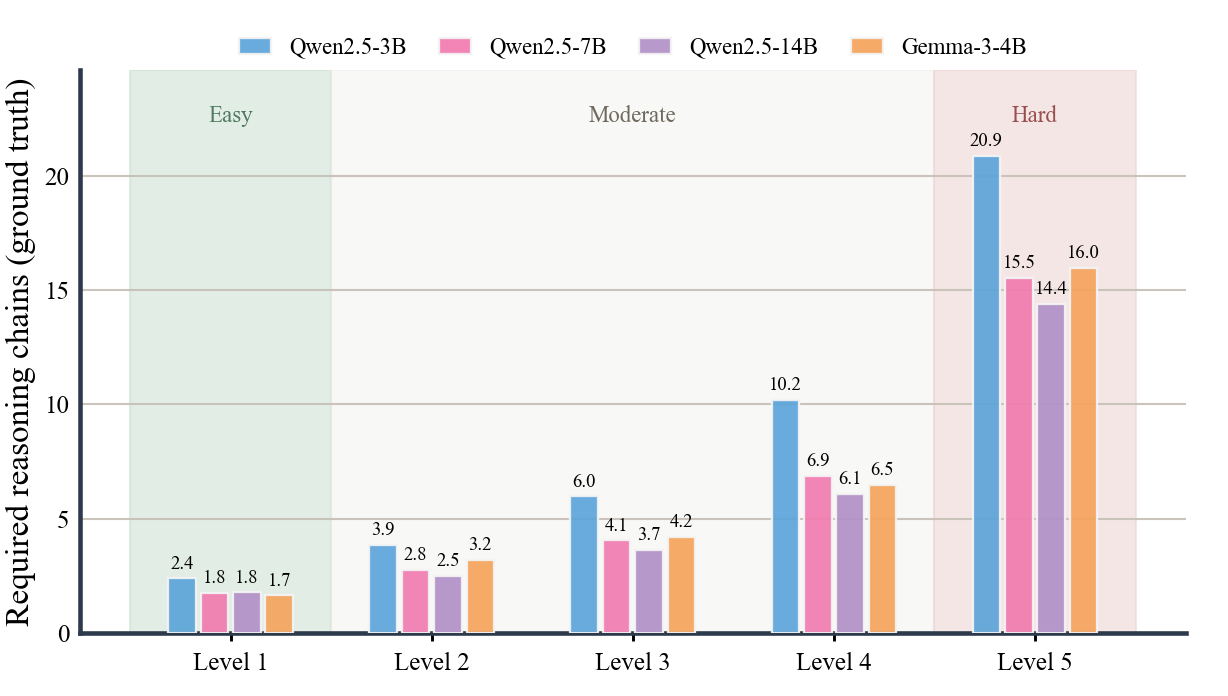}
        \caption{Required reasoning chains across difficulty levels.}
        \label{fig: visualize_sc_consensus_and_answer_diversity_math_panel_left}
    \end{subfigure}
    \hfill
    \begin{subfigure}[b]{0.38\linewidth}
        \centering
        \includegraphics[width=\linewidth]{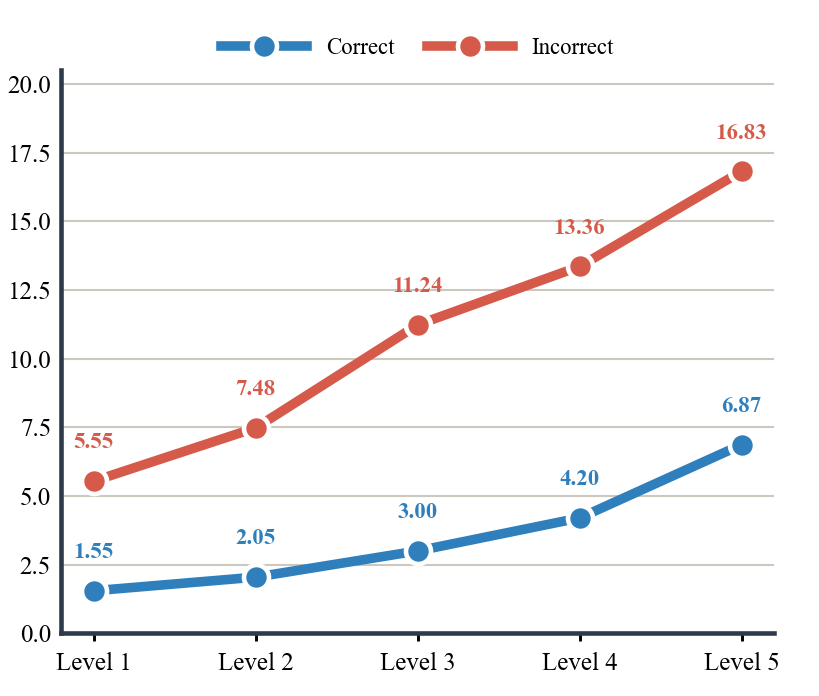}
        \caption{Answer diversity across difficulty levels.}
        \label{fig: visualize_sc_consensus_and_answer_diversity_math_panel_right}
    \end{subfigure}
    \caption{\textbf{Empirical motivation for adaptive Self-Consistency.}
    (a): As problem difficulty increases, the number of reasoning chains required to reach the correct answer also increases.
    (b): Answer diversity also increases alongside difficulty, with incorrect answers showing particularly higher diversity, suggesting that uncertainty becomes greater for more difficult problems.}
    \label{fig:overall}
\end{figure*}

\section{Motivation}
\label{sec: Motivation}

In this section, we empirically show that the amount of reasoning required varies across problems and that problem difficulty can be represented as a continuous signal. To this end, we use the MATH training dataset~\citep{math}, where difficulty levels are explicitly defined, and generate 40 reasoning chains for each problem with four instruction-tuned LLMs: Qwen2.5-Instruct (3B, 7B, 14B)~\cite{qwen2.5}, and Gemma-3-4B-it~\cite{gemma_3}. We then analyze the generated chains from multiple perspectives.

\paragraph{Do all questions require the same number of reasoning paths?}
SC assigns a fixed sampling budget to all inputs, but the number of reasoning paths required to reach the correct answer may vary depending on problem difficulty. To verify this, we sequentially accumulate the generated chains and measure the minimum number of chains at which the majority-voting result first matches the correct answer. In addition, when the correct answer is not reached even after using all chains, we assign the maximum number of chains so that failure cases are reflected.

As shown in Figure~\ref{fig: visualize_sc_consensus_and_answer_diversity_math_panel_left}, across all models, the average number of reasoning chains required to reach the correct answer tends to increase as the difficulty level of the problem increases. In particular, in the high-difficulty range (Level 5), the required number of chains increases sharply, and this trend appears consistently across different models.

These results suggest that the amount of reasoning required varies across problems and that reasoning resources need to be flexibly adjusted according to problem difficulty.

\paragraph{Can problem difficulty be modeled as a continuous signal?}
From the preceding analysis, we confirm that the number of reasoning paths required by a problem gradually increases in proportion to its difficulty. In other words, reasoning resources should be allocated in proportion to difficulty, and accurately identifying problem difficulty is therefore important.

Most existing studies distinguish difficulty coarsely, such as easy or hard, and apply different reasoning strategies accordingly~\citep{dsc, llm_already_knows, diffadapt}. However, since the degree of difficulty perceived by the model can differ even among different problems at the same level, discrete difficulty categorization cannot precisely reflect the amount of reasoning required for each problem. Given this, can the difficulty perceived by the model be quantified as a continuous signal and represented in a more fine-grained manner?

\begin{figure*}[!t]
  \includegraphics[width=\linewidth]{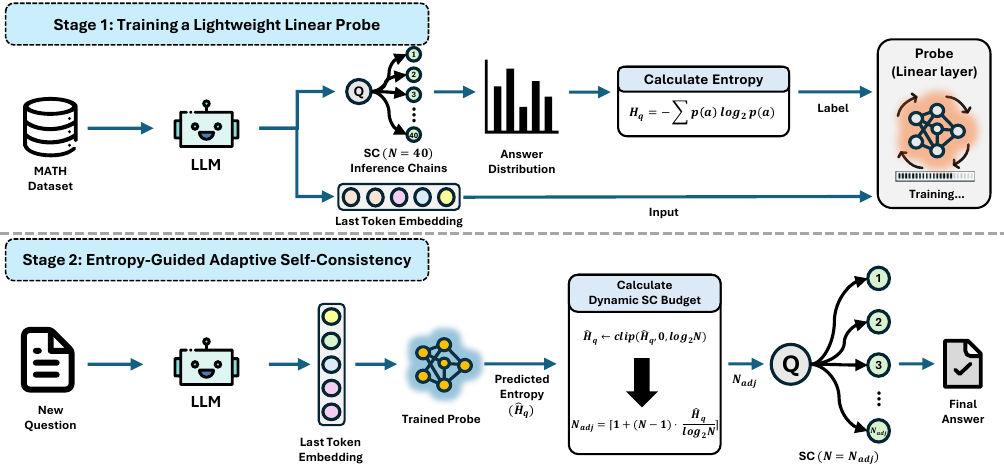}
  \caption{\textbf{Overall framework of FSC.} 
  In Stage 1, FSC trains a lightweight probe to predict output entropy as a continuous difficulty signal. 
  In Stage 2, the predicted entropy is mapped to an adaptive sampling budget, which determines the number of reasoning chains used for majority voting.}
  \label{fig: fsc_framework}
\end{figure*}

To this end, we focus on the entropy of the model's output distribution as such a signal. As shown in Figure~\ref{fig: visualize_sc_consensus_and_answer_diversity_math_panel_right}, the number of unique answers tends to increase as difficulty increases, and incorrect cases in particular exhibit higher diversity than correct cases. This trend appears consistently across different models (details in Appendix~\ref{sec:appendix motivation}), suggesting that the entropy of the response distribution reflects an intrinsic signal of the problem difficulty perceived by the model. Based on this observation, we propose \textbf{F}lexible \textbf{S}elf-\textbf{C}onsistency (\textbf{FSC}), which adaptively adjusts the sampling budget using output entropy.

\section{Methodology}
\label{sec: Flexible Self-Consistency}

As illustrated in Figure~\ref{fig: fsc_framework}, FSC consists of two stages. First, a lightweight linear probe is trained to predict the output entropy based on the last-token embedding of the LLM given an input prompt. This design is motivated by prior studies showing that embeddings produced by LLMs can reflect intrinsic properties of prompts~\citep{last_embed_literary, last_embed_instruct, last_embed_safety, probing_difficulty}. Next, the trained probe predicts the entropy of a new input question, which is then used to allocate an appropriate sampling budget for majority voting.

\subsection{Training a Lightweight Linear Probe}
\label{subsec: Training a Lightweight Linear Probe}

\paragraph{Data Collection for Probe.}
\label{parag: Data Collection for Probe.}

For each input question $q \in D$, we first use an LLM to generate a set of $N$ reasoning trajectories, denoted as $\mathcal{R}_q=\{r_i\}_{i=1}^{N}$. We then extract the final answers from these trajectories to construct the answer set $\mathcal{A}_q=\{a_i\}_{i=1}^{N}$.

Second, let $\mathcal{U}_q$ denote the set of unique answers in $\mathcal{A}_q$. Based on the relative frequency of each answer $a \in \mathcal{U}_q$, we define a probability distribution $p_q(a)$ and compute the output entropy $H_q$. Specifically, let $c_q(a)$ denote the occurrence count of answer $a$. Then, $p_q(a)$ and $H_q$ are computed as follows:
\begin{equation}
\begin{gathered}
p_q(a) = \frac{c_q(a)}{N}, \\
H_q = - \sum_{a \in \mathcal{U}_q} p_q(a)\log_2 p_q(a)
\end{gathered}
\end{equation}
The resulting entropy $H_q$ is used as the entropy label for input question $q$, yielding a synthetic dataset of the form $\tilde{D}=\{(q, H_q) \mid q \in D\}$.

Unlike prior studies that train probes using discrete difficulty labels~\citep{probing_difficulty, diffadapt}, our entropy-based labeling provides a continuous supervision signal that can capture problem difficulty in a more fine-grained manner.

\paragraph{Probe Learning.}
\label{parag: Probe Learning}

Using the constructed dataset $\tilde{D}$, the probe is trained to predict the output entropy from the last-token embedding of the LLM for each input question. Specifically, the probe is optimized using the mean squared error loss between the predicted entropy $\hat{H}_q$ and the labeled entropy $H_q$:

\begin{equation}
\mathcal{L}
=
\mathbb{E}_{(q,H_q)\sim\tilde{D}}
\left[
\left(H_q-\hat{H}_q\right)^2
\right]
\end{equation}
To model continuous and fine-grained difficulty signals, the probe is implemented as a linear regression model without any non-linear activation function.

\subsection{Entropy-Guided Adaptive Self-Consistency}
\label{subsec: Entropy-Guided Adaptive Self-Consistency}

\paragraph{Entropy Estimation.}
\label{parag: Entropy Estimation}

Following the training procedure, the probe takes the last-token embedding of the LLM for a new input question and predicts its output entropy.

In our setting, the answer distribution is constructed from at most $N$ reasoning paths, meaning that the number of unique answers cannot exceed $N$. Accordingly, the entropy of the answer distribution is theoretically bounded within the range $[0, \log_2 N]$~\citep{shannon_entropy}. However, since the probe predicts a continuous value $\hat{H}_q$, the predicted entropy may fall outside this range. Therefore, similar to prior studies~\citep{ppo, grpo, adaptthink}, we apply clipping for stability as follows:

\begin{equation}
\hat{H}_q \leftarrow \operatorname{clip}(\hat{H}_q,\ 0,\ \log_2 N)
\end{equation}

\paragraph{Adjusting Sampling Budget.}
\label{parag: Adjusting Sampling Budget}
The entropy predicted by the probe serves as the basis for dynamically adjusting the sampling budget for each question. However, directly using the predicted entropy makes it difficult to maintain a consistent interpretation across different budget settings. Therefore, we normalize the entropy into a relative difficulty score within the range $[0,1]$. We then determine the appropriate sampling budget $N_{\text{adj}}$ for an input question $q$ in proportion to the normalized entropy as follows:

\begin{equation}
N_{\text{adj}} = \left\lceil 1 + (N-1) \cdot \frac{\hat{H}_q}{\log_2 N} \right\rceil
\end{equation}

This budget allocation strategy enables adaptive reasoning proportional to the continuous difficulty of each question by assigning a minimum of one reasoning path to easy questions with low entropy and up to $N$ reasoning paths to difficult questions with high entropy.

\begin{table*}[t]
\centering
\small
\renewcommand{\arraystretch}{1.2}
\setlength{\tabcolsep}{2.2pt} 
\resizebox{\textwidth}{!}{
\begin{tabular}{c cc cc cc cc cc}
\toprule
\multirow{2}{*}{\textbf{Methods}} 
& \multicolumn{2}{c}{\textbf{MATH500}}
& \multicolumn{2}{c}{\textbf{AMC23}}
& \multicolumn{2}{c}{\textbf{AIME2024}} 
& \multicolumn{2}{c}{\textbf{AIME2025}}
& \multicolumn{2}{c}{\textbf{GPQA-D}} \\
\cmidrule(lr){2-3} 
\cmidrule(lr){4-5} 
\cmidrule(lr){6-7} 
\cmidrule(lr){8-9}
\cmidrule(lr){10-11}
& Acc.↑ & Tok.↓ 
& Acc.↑ & Tok.↓ 
& Acc.↑ & Tok.↓ 
& Acc.↑ & Tok.↓
& Acc.↑ & Tok.↓ \\
\midrule

\multicolumn{11}{c}{\cellcolor{Oursgray}\textbf{Qwen2.5-3B}} \\ \midrule

SC
& 74.6 & 24.4 (\textcolor{gray}{0.0\%})
& 52.5 & 34.4 (\textcolor{gray}{0.0\%})
& 16.7 & 43.7 (\textcolor{gray}{0.0\%})
& 3.3  & 36.8 (\textcolor{gray}{0.0\%})
& 30.3 & 27.8 (\textcolor{gray}{0.0\%}) \\
\hdashline

AC
& 74.8 & 13.3 (\textcolor{DeltaGreen}{-45.5\%})
& 52.5 & 23.6 (\textcolor{DeltaGreen}{-31.4\%})
& 16.7 & 43.6 (\textcolor{DeltaGreen}{-0.2\%})
& 3.3  & 42.6 (\textcolor{blue}{+15.8\%})
& 30.3 & 23.2 (\textcolor{DeltaGreen}{-16.2\%}) \\

ESC
& 74.6 & 15.6 (\textcolor{DeltaGreen}{-36.1\%})
& 52.5 & 30.6 (\textcolor{DeltaGreen}{-11.0\%})
& 16.7 & 44.0 (\textcolor{blue}{+0.7\%})
& 3.3  & 38.0 (\textcolor{blue}{+3.3\%})
& 29.7 & 23.7 (\textcolor{DeltaGreen}{-14.7\%}) \\

DSC
& 74.6 & 12.9 (\textcolor{DeltaGreen}{-47.1\%})
& 52.5 & 22.9 (\textcolor{DeltaGreen}{-33.4\%})
& 16.7 & 41.3 (\textcolor{DeltaGreen}{-5.5\%})
& 3.3  & 36.8 (\textcolor{DeltaGreen}{0.0\%})
& 30.8 & 22.0 (\textcolor{DeltaGreen}{-20.8\%}) \\

\rowcolor{PastelBlue}
\textbf{FSC (Ours)}
& 74.6 & \textbf{10.4 (\textcolor{red}{-57.4\%})}
& 52.5 & \textbf{19.0 (\textcolor{red}{-44.8\%})}
& 13.3 & \textbf{33.2 (\textcolor{red}{-24.0\%})}
& 3.3  & \textbf{28.9 (\textcolor{red}{-21.5\%})}
& 31.8 & \textbf{21.9 (\textcolor{red}{-21.1\%})} \\

\specialrule{.12em}{.4em}{.4em}

\multicolumn{11}{c}{\cellcolor{Oursgray}\textbf{Qwen2.5-7B}} \\ \midrule

SC
& 80.6 & 24.2 (\textcolor{gray}{0.0\%})
& 62.5 & 34.7 (\textcolor{gray}{0.0\%})
& 13.3 & 42.8 (\textcolor{gray}{0.0\%})
& 20.0 & 38.4 (\textcolor{gray}{0.0\%})
& 36.9 & 22.1 (\textcolor{gray}{0.0\%}) \\
\hdashline

AC
& 80.6 & 10.1 (\textcolor{DeltaGreen}{-58.3\%})
& 62.5 & 19.5 (\textcolor{DeltaGreen}{-43.8\%})
& 13.3 & 38.6 (\textcolor{DeltaGreen}{-9.8\%})
& 20.0 & 41.0 (\textcolor{blue}{+6.8\%})
& 36.4 & 14.9 (\textcolor{DeltaGreen}{-32.7\%}) \\

ESC
& 80.6 & 12.2 (\textcolor{DeltaGreen}{-49.6\%})
& 62.5 & 26.3 (\textcolor{DeltaGreen}{-24.2\%})
& 13.3 & 40.8 (\textcolor{DeltaGreen}{-4.7\%})
& 20.0 & 38.4 (\textcolor{DeltaGreen}{0.0\%})
& 37.4 & 15.7 (\textcolor{DeltaGreen}{-28.9\%}) \\

DSC
& 80.6 & 10.0 (\textcolor{DeltaGreen}{-58.7\%})
& 62.5 & 20.0 (\textcolor{DeltaGreen}{-42.4\%})
& 13.3 & 36.6 (\textcolor{DeltaGreen}{-14.5\%})
& 20.0 & 36.2 (\textcolor{DeltaGreen}{-5.7\%})
& 36.9 & 13.2 (\textcolor{DeltaGreen}{-40.1\%}) \\

\rowcolor{PastelBlue}
\textbf{FSC (Ours)}
& 80.6 & \textbf{7.0 (\textcolor{red}{-71.1\%})}
& 65.0 & \textbf{15.1 (\textcolor{red}{-56.5\%})}
& 16.7 & \textbf{27.5 (\textcolor{red}{-35.7\%})}
& 16.7 & \textbf{27.7 (\textcolor{red}{-27.9\%})}
& 36.4 & \textbf{10.6 (\textcolor{red}{-51.9\%})} \\

\specialrule{.12em}{.4em}{.4em}

\multicolumn{11}{c}{\cellcolor{Oursgray}\textbf{Qwen2.5-14B}} \\ \midrule

SC
& 81.6 & 24.5 (\textcolor{gray}{0.0\%})
& 70.0 & 35.3 (\textcolor{gray}{0.0\%})
& 20.0 & 42.4 (\textcolor{gray}{0.0\%})
& 23.3 & 39.5 (\textcolor{gray}{0.0\%})
& 46.2 & 23.6 (\textcolor{gray}{0.0\%}) \\
\hdashline

AC
& 81.8 & 9.0 (\textcolor{DeltaGreen}{-63.3\%})
& 70.0 & 16.9 (\textcolor{DeltaGreen}{-52.1\%})
& 20.0 & 37.1 (\textcolor{DeltaGreen}{-12.5\%})
& 23.3 & 32.6 (\textcolor{DeltaGreen}{-17.5\%})
& 45.6 & 14.6 (\textcolor{DeltaGreen}{-38.1\%}) \\

ESC
& 81.6 & 10.7 (\textcolor{DeltaGreen}{-56.3\%})
& 70.0 & 23.2 (\textcolor{DeltaGreen}{-34.3\%})
& 20.0 & 39.1 (\textcolor{DeltaGreen}{-7.8\%})
& 23.3 & 32.2 (\textcolor{DeltaGreen}{-18.5\%})
& 46.7 & 15.3 (\textcolor{DeltaGreen}{-35.0\%}) \\

DSC
& 81.6 & 8.9 (\textcolor{DeltaGreen}{-63.7\%})
& 70.0 & 18.1 (\textcolor{DeltaGreen}{-48.7\%})
& 20.0 & 35.4 (\textcolor{DeltaGreen}{-16.5\%})
& 23.3 & 27.4 (\textcolor{DeltaGreen}{-30.6\%})
& 46.2 & 13.4 (\textcolor{DeltaGreen}{-43.1\%}) \\

\rowcolor{PastelBlue}
\textbf{FSC (Ours)}
& 82.2 & \textbf{6.1 (\textcolor{red}{-75.1\%})}
& 70.0 & \textbf{14.0 (\textcolor{red}{-60.3\%})}
& 20.0 & \textbf{25.9 (\textcolor{red}{-38.9\%})}
& 23.3 & \textbf{26.3 (\textcolor{red}{-33.4\%})}
& 46.7 & \textbf{7.5 (\textcolor{red}{-68.2\%})} \\

\specialrule{.12em}{.4em}{.4em}

\multicolumn{11}{c}{\cellcolor{Oursgray}\textbf{Gemma-3-4B}} \\ \midrule

SC
& 79.2 & 36.5 (\textcolor{gray}{0.0\%})
& 50.0 & 52.5 (\textcolor{gray}{0.0\%})
& 13.3 & 76.0 (\textcolor{gray}{0.0\%})
& 13.3 & 62.6 (\textcolor{gray}{0.0\%})
& 28.2 & 38.6 (\textcolor{gray}{0.0\%}) \\
\hdashline

AC
& 79.2 & 15.7 (\textcolor{DeltaGreen}{-57.0\%})
& 50.0 & 34.9 (\textcolor{DeltaGreen}{-33.5\%})
& 13.3 & 60.4 (\textcolor{DeltaGreen}{-20.5\%})
& 13.3 & 53.1 (\textcolor{DeltaGreen}{-15.2\%})
& 27.7 & 29.9 (\textcolor{DeltaGreen}{-22.5\%}) \\

ESC
& 79.2 & 19.7 (\textcolor{DeltaGreen}{-46.0\%})
& 50.0 & 41.0 (\textcolor{DeltaGreen}{-21.9\%})
& 13.3 & 71.7 (\textcolor{DeltaGreen}{-5.7\%})
& 13.3 & 52.7 (\textcolor{DeltaGreen}{-15.8\%})
& 27.7 & 33.2 (\textcolor{DeltaGreen}{-13.9\%}) \\

DSC
& 79.2 & 16.5 (\textcolor{DeltaGreen}{-54.8\%})
& 50.0 & 35.6 (\textcolor{DeltaGreen}{-32.2\%})
& 13.3 & 62.9 (\textcolor{DeltaGreen}{-17.2\%})
& 13.3 & 50.6 (\textcolor{DeltaGreen}{-19.2\%})
& 28.2 & 31.0 (\textcolor{DeltaGreen}{-19.7\%}) \\

\rowcolor{PastelBlue}
\textbf{FSC (Ours)}
& 78.8 & \textbf{8.5 (\textcolor{red}{-76.7\%})}
& 52.5 & \textbf{18.5 (\textcolor{red}{-64.8\%})}
& 13.3 & \textbf{29.4 (\textcolor{red}{-61.3\%})}
& 16.7 & \textbf{30.9 (\textcolor{red}{-50.6\%})}
& 26.2 & \textbf{9.7 (\textcolor{red}{-74.8\%})} \\

\bottomrule
\end{tabular}
}
\caption{\textbf{Main experimental results across models and benchmarks.}
Accuracies are reported in \% (rounded to one decimal).
Tok. denotes total tokens (input + output), reported in $10^3$ tokens. Values in parentheses denote the percentage change in token usage relative to SC for each model--dataset combination.}
\label{tab: Table1}
\end{table*}

\section{Experiments}
\label{sec: Experiments}

\subsection{Setup}
\label{subsec: Setup}

\paragraph{Datasets.}
To train the probe, we use the MATH~\citep{math} training dataset consisting of 7,500 instances. For evaluation, we use MATH500~\citep{math500}, AMC23~\citep{amc23}, AIME2024~\citep{aime24}, AIME2025~\citep{aime25}, and GPQA-Diamond~\citep{gpqa}. MATH500 contains diverse mathematical reasoning problems, while AMC and AIME consist of challenging mathematics competition problems. GPQA-Diamond is a benchmark designed to evaluate graduate-level STEM reasoning ability.

\paragraph{Models.}
To cover a diverse range of model families and scales, we use instruction-tuned language models that have demonstrated strong reasoning capabilities, including Qwen2.5-Instruct (3B, 7B, and 14B) and Gemma-3-4B-it. All experiments are conducted under a zero-shot prompting setting, and detailed prompts are provided in Appendix~\ref{sec:appendix Full Prompt}.

\paragraph{Baselines and Metrics.}
To evaluate the performance of FSC, we compare it with the following SC-based methods:
\begin{itemize}
    \item \textbf{SC} \citep{sc}: A method that generates multiple reasoning chains for the same problem and selects the final answer by applying majority voting over the generated answers.
    \item \textbf{AC} \citep{ac}: A method that sequentially generates reasoning chains, measures the consistency of intermediate results, and terminates sampling early when the confidence exceeds a predefined threshold.
    \item \textbf{ESC} \citep{esc}: A method that generates reasoning chains progressively and stops additional sampling when the generated answers sufficiently converge within a fixed window.
    \item \textbf{DSC} \citep{dsc}: A difficulty-aware method that estimates the difficulty of input problems discretely based on the LLM's self-assessment. It allocates a single reasoning path to easy problems, while adaptively increasing the number of reasoning paths for hard problems through additional sampling and sample size pre-allocation.
\end{itemize}

For fair comparison, the maximum sampling budget of SC is fixed to $N=40$, and the resulting reasoning paths are used as a shared pool so that all methods select from the same set of reasoning trajectories. Performance is evaluated using accuracy and token usage (input + output). Additional implementation details and experimental settings are provided in Appendix~\ref{sec:appendix detailed experimental setup}.

\subsection{Results}
\label{subsec: Results}
Table~\ref{tab: Table1} presents the comparison between FSC and existing methods across various models and benchmarks. For the Qwen family series, FSC and all baselines generally exhibit improved token efficiency across all benchmarks as the model size increases. However, this trend is more pronounced for FSC, which achieves the lowest token usage in most settings while maintaining accuracy comparable to existing baselines. In particular, FSC consistently reduces token consumption across all model--dataset combinations, achieving up to a 76.7\% reduction compared to SC.

In contrast, AC and ESC occasionally consume even more tokens than SC in certain settings. For example, on AIME2025 with Qwen2.5-3B, AC increases token usage by +15.8\% compared to SC, while Qwen2.5-7B shows a +6.8\% increase.

Furthermore, compared to DSC, FSC demonstrates higher token efficiency while maintaining comparable accuracy in most settings. We attribute this improvement to FSC's use of continuous signals derived from output entropy, which enables more fine-grained resource allocation than DSC's discrete difficulty estimation. Additional analysis is provided in \S\ref{subsec: Comparison of Inference Path Assignment Distributions by Difficulty Level}.

Overall, FSC effectively balances accuracy and efficiency, demonstrating a more stable and efficient test-time scaling strategy compared to existing approaches. Additional qualitative case studies are provided in Appendix~\ref{sec: case_study}, highlighting how FSC allocates inference paths more efficiently than existing methods across problem difficulty.

\section{Analysis}
\label{sec: Analysis}

\subsection{Can the Probe Reflect Problem Difficulty via Entropy?}
\label{subsec: Can the Probe Reflect Problem Difficulty via Entropy}
\begin{figure}[h]
  \includegraphics[width=\columnwidth]{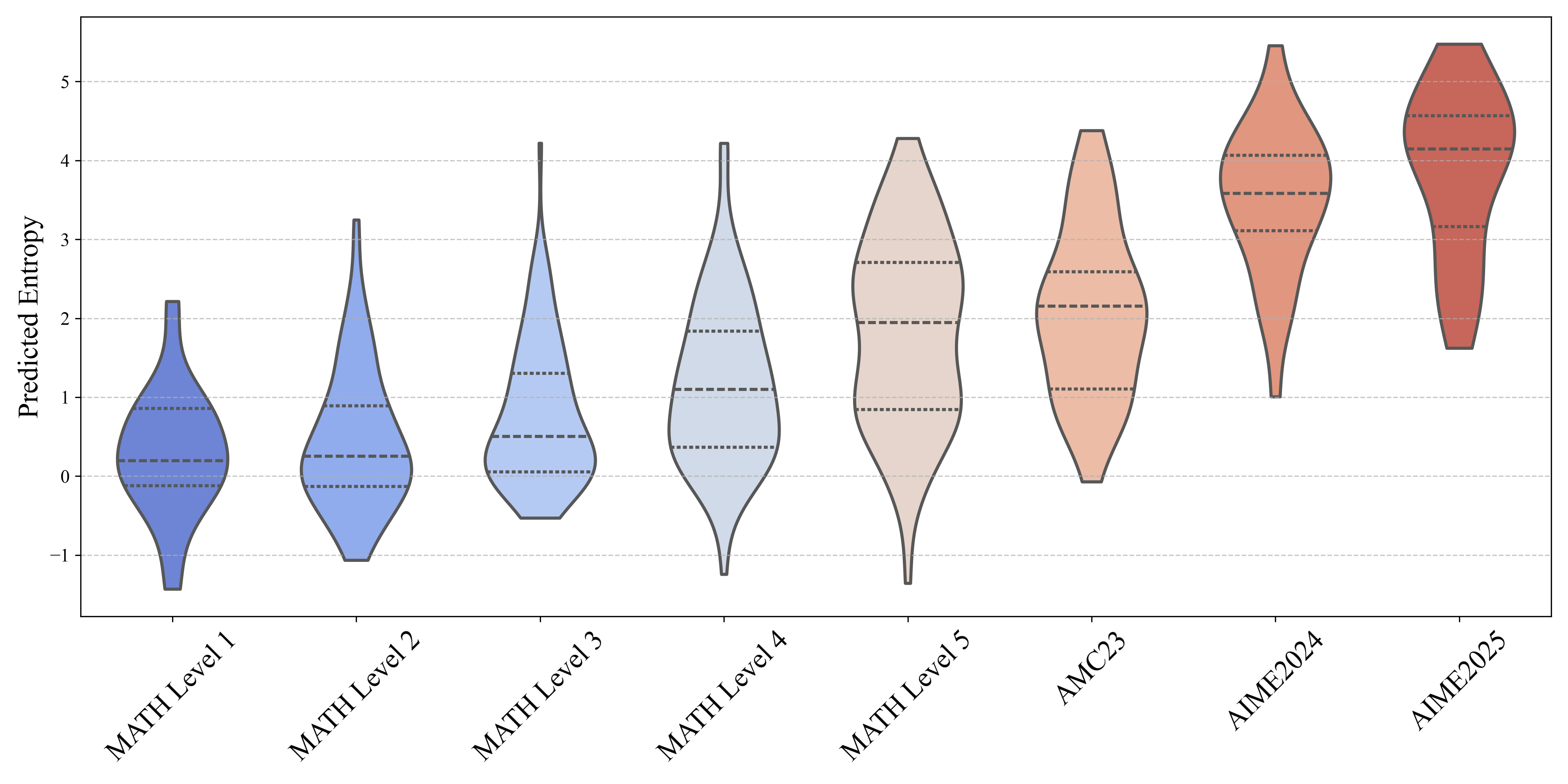}
  \caption{Distribution of predicted entropy by probe across problem difficulty levels and STEM benchmarks using Qwen2.5-7B.}
  \label{fig:entropy_violin_qwen2.5_7b}
\end{figure}
Figure~\ref{fig:entropy_violin_qwen2.5_7b} illustrates how the entropy predicted by the trained probe corresponds to problem difficulty. In this analysis, we use four mathematical reasoning benchmarks: MATH500, AMC23, AIME2024, and AIME2025.

\begin{figure*}[t]
  \includegraphics[width=\linewidth]{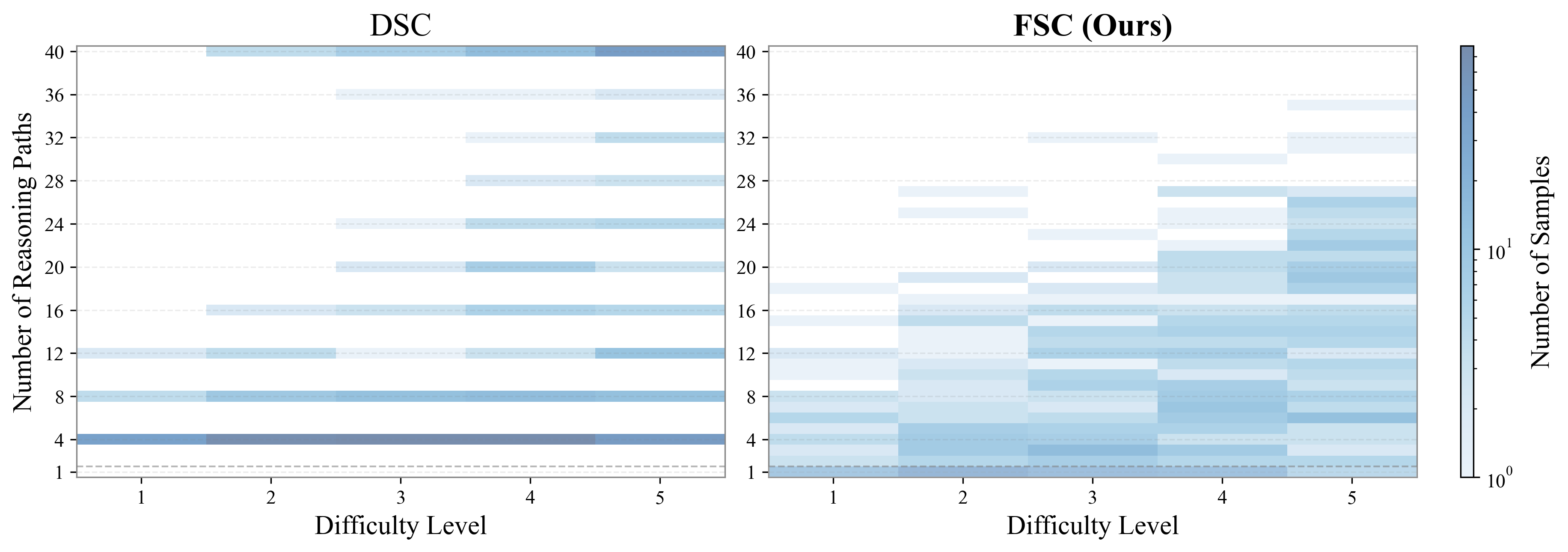}
  \caption{Comparison of inference path allocation distributions by difficulty level between DSC and FSC on MATH500 using Qwen2.5-7B.}
  \label{fig:path_allocation_qwen2.5_7b}
\end{figure*}

\begin{figure*}[t]
  \includegraphics[width=\linewidth]{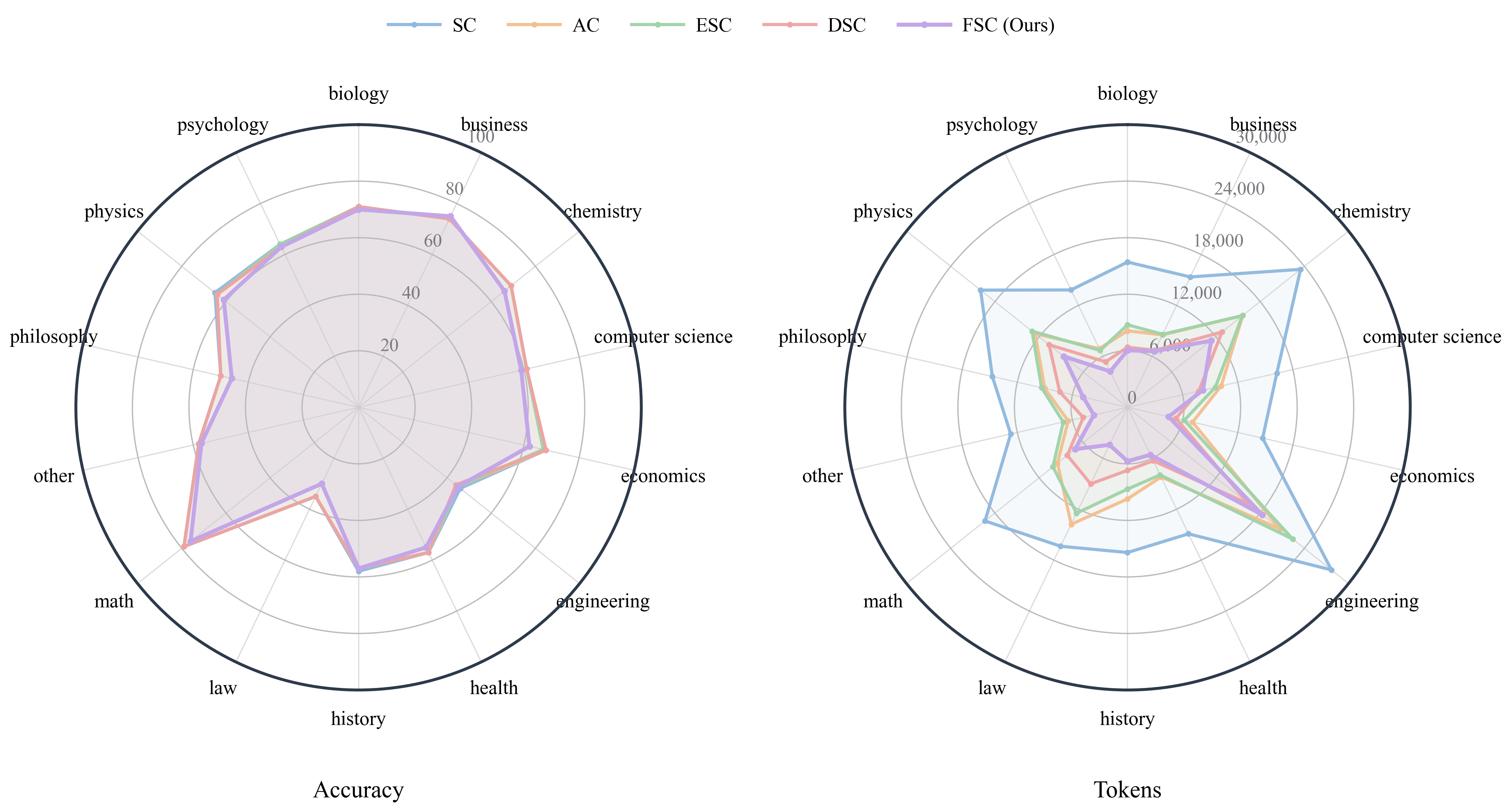}
    \caption{\textbf{Token efficiency comparison on MMLU-Pro with Qwen2.5-7B.} FSC achieves the highest efficiency in most domains and demonstrates stable performance across diverse distributional settings.}
    \label{fig: mmlu_pro_category_radar_qwen2.5-7b-instruct}
\end{figure*}

Overall, the predicted entropy tends to increase as problem difficulty increases. First, as observed in MATH500, where explicit difficulty labels are provided, the predicted entropy gradually increases with higher difficulty levels. Furthermore, the entropy distributions of AMC23, AIME2024, and AIME2025 are overall higher than those of the highest difficulty level in MATH500 (Level 5), consistent with the fact that these benchmarks are generally considered more challenging~\citep{analyis_prob_diff1, analysis_prob_diff2}.

Similar trends are observed across other models, and the corresponding results are presented in Figure~\ref{fig: entropy_violin_all}. These results suggest that the trained probe does not merely overfit to a specific dataset, but instead learns a generalized continuous difficulty signal that reflects the absolute difficulty of problems. 

\subsection{Comparison of Inference Path Assignment Distributions by Difficulty Level}
\label{subsec: Comparison of Inference Path Assignment Distributions by Difficulty Level}
Figure \ref{fig:path_allocation_qwen2.5_7b} shows the difference in the distribution of reasoning paths allocated by DSC and FSC for each difficulty level on MATH500. For both DSC and FSC, the distribution progressively shifts toward larger numbers of reasoning paths as the difficulty increases. However, the two methods show a clear difference in how the number of reasoning paths is distributed.

Specifically, DSC tends to concentrate resource allocation within certain ranges, whereas FSC forms a more continuous distribution over a broader range without being biased toward specific values. This difference stems from the difficulty estimation and resource allocation mechanisms of each method. DSC estimates problem difficulty discretely, assigning only a single reasoning path to easy problems and performing adaptive sampling for difficult problems until sufficient agreement among responses is achieved. In contrast, FSC estimates difficulty continuously based on the output entropy of the input question and flexibly allocates reasoning resources in proportion to it. These distributional characteristics are not limited to a specific model, and similar trends are consistently observed across various models. Detailed results are presented in Figure~\ref{fig: path_allocation_all}. This suggests that FSC more precisely reflects the intrinsic difficulty of each input question, enabling more efficient and flexible resource allocation.

\subsection{Generalization under Distribution Shift}
\label{subsec: Generalization under Distribution Shift}
While \S\ref{subsec: Results} evaluates STEM-oriented reasoning ability using MATH500, AMC23, AIME2024, AIME2025, and GPQA-Diamond, we further assess the Out-of-Distribution (OOD) performance of FSC using MMLU-Pro~\citep{mmlu_pro}, which covers a broad range of academic disciplines. 
Since MMLU-Pro is a large-scale benchmark spanning diverse domains, we sample a subset for efficient evaluation. Specifically, we randomly sample 100 questions from each of the 14 categories using a fixed random seed of 42, resulting in a total of 1,400 evaluation samples.

As shown in Figure~\ref{fig: mmlu_pro_category_radar_qwen2.5-7b-instruct}, FSC achieves the highest efficiency in most academic domains and shows consistent performance across diverse domains. 
This trend is consistently observed across other models as well, as shown in Figure~\ref{fig: mmlu_pro_category_radar_all}, suggesting that FSC is not dependent on a specific model or parameter scale. Overall, these results show that FSC stably maintains high token efficiency across diverse domains and effectively generalizes under distribution shift as a test-time scaling method.

\section{Related Works}
\label{sec: Related Works}

\paragraph{Efficient Self-Consistency Variants.}
Self-Consistency (SC) shows strong performance across diverse reasoning tasks by sampling multiple reasoning trajectories and determining the final response through majority voting~\citep{sc}. Nevertheless, the increased token consumption caused by generating multiple reasoning paths limits the scalability of SC, and subsequent studies have proposed various methods to improve its efficiency. For example, AC~\citep{ac} adopts a sequential sampling strategy, while ESC~\citep{esc} uses small-batch sampling; both reduce token consumption by stopping sampling early according to certain criteria, such as consistency within the answer distribution. DSC~\citep{dsc} estimates the difficulty of a question and samples proportionally to the estimated difficulty, showing that inference cost can be reduced more substantially than in prior approaches. However, since such difficulty-based efficient sampling strategies require accurate difficulty estimation as a prerequisite, various studies on difficulty estimation for LLMs have recently been actively explored.

\paragraph{Difficulty Estimation.}
Research on difficulty estimation for LLMs has developed mainly along two directions: external estimation and internal estimation. External estimation methods either make LLMs explicitly judge difficulty or implicitly estimate difficulty through fine-tuning~\citep{dsc, adaptthink, diff_aware_cot}. In contrast, internal estimation methods leverage the model's internal signals to more precisely estimate the difficulty perceived by the model. For example, \citet{llm_already_knows} shows that difficulty can be effectively predicted through a value function without token generation, while \citet{probing_difficulty} demonstrates that a linear probe can effectively predict the difficulty of an input question.

However, existing methods still often treat difficulty as discrete categories or coarse levels, such as easy and hard, and thus fail to sufficiently reflect the difficulty actually perceived by the model. To address this limitation, FSC represents difficulty as a continuous signal and uses it to guide the allocation of reasoning resources, moving beyond coarse difficulty categories.

\section{Conclusion}
\label{sec: Conclusion}
In this paper, we show that output entropy can serve as a continuous difficulty signal that replaces conventional discrete difficulty categorization. We further propose FSC, which predicts entropy using a lightweight linear probe and dynamically allocates the sampling budget accordingly. Across various benchmarks and models, FSC substantially reduces token usage while maintaining accuracy comparable to existing baselines. These results suggest that FSC effectively improves the efficiency and practicality of self-consistency while preserving its performance benefits by flexibly adjusting computational resources according to problem difficulty.

\section*{Limitations}
\label{sec: Limitations}
Despite the strong performance of FSC, several limitations remain. First, due to resource constraints, we could not conduct experiments on models larger than 14B. In particular, since the entropy predicted by the probe may vary depending on model size and family, further validation is required. Second, FSC requires the hidden representation of the last token of the input question to train the probe, making it difficult to apply to proprietary models such as GPT. Therefore, developing a more scalable efficient self-consistency method that overcomes these constraints and can be applied across diverse model environments remains an important direction for subsequent work. Finally, since the probe in this work was trained only on mathematical datasets, training probe on more diverse datasets is expected to further improve generalization and robustness across a wider range of tasks and domains.

\section*{Acknowledgements}

This work was supported by the National Research Foundation of Korea (NRF) grant funded by the Korea government (MSIT) (RS-2025-00553041, Enhancement of Rational and Emotional Intelligence of Large Language Models for Implementing Dependable Conversational Agents, Contribution Rate: 50\%).
This research was supported by the Culture, Sports and Tourism R\&D Program through the Korea Creative Content Agency grant funded by the Ministry of Culture, Sports and Tourism in 2026 (Project Name: Development of an AI Agent Integrating Korean Language Knowledge for Personalized Language Consultation Services, Project Number: RS-2026-25506607, Contribution Rate: 50\%).

\bibliography{custom}

@article{scaling_laws,
  title={Scaling laws for neural language models},
  author={Kaplan, Jared and McCandlish, Sam and Henighan, Tom and Brown, Tom B and Chess, Benjamin and Child, Rewon and Gray, Scott and Radford, Alec and Wu, Jeffrey and Amodei, Dario},
  journal={arXiv preprint arXiv:2001.08361},
  year={2020}
}

@article{chinchilla,
  title={Training compute-optimal large language models},
  author={Hoffmann, Jordan and Borgeaud, Sebastian and Mensch, Arthur and Buchatskaya, Elena and Cai, Trevor and Rutherford, Eliza and Casas, DDL and Hendricks, Lisa Anne and Welbl, Johannes and Clark, Aidan and others},
  journal={arXiv preprint arXiv:2203.15556},
  volume={10},
  year={2022}
}

@article{llm_survey,
  title={A survey of large language models},
  author={Zhao, Wayne Xin and Zhou, Kun and Li, Junyi and Tang, Tianyi and Wang, Xiaolei and Hou, Yupeng and Min, Yingqian and Zhang, Beichen and Zhang, Junjie and Dong, Zican and others},
  journal={arXiv preprint arXiv:2303.18223},
  volume={1},
  number={2},
  pages={1--124},
  year={2023}
}

@inproceedings{geunyeong,
  title={Watch Your Step: A Fine-Grained Evaluation Framework for Multi-hop Knowledge Editing in Large Language Models},
  author={Jeong, Geunyeong and Sun, Juoh and Kim, Harksoo},
  booktitle={Proceedings of the 34th ACM International Conference on Information and Knowledge Management},
  pages={4842--4846},
  year={2025}
}

@article{byungkook,
  title={Interaction-Grounded Semantic Graph Refinement for LLM-Based Recommendation},
  author={Jeong, Wooseok and Kim, Young-Jin and Koo, Hae-Yoon and Seo, Jimyeung and Choi, Jinho and Oh, Byungkook},
  journal={IEEE Access},
  volume={13},
  pages={194229--194244},
  year={2025},
  publisher={IEEE}
}

@article{tts_survey,
  title={A survey on test-time scaling in large language models: What, how, where, and how well?},
  author={Zhang, Qiyuan and Lyu, Fuyuan and Sun, Zexu and Wang, Lei and Zhang, Weixu and Hua, Wenyue and Wu, Haolun and Guo, Zhihan and Wang, Yufei and Muennighoff, Niklas and others},
  journal={arXiv preprint arXiv:2503.24235},
  year={2025}
}

@article{tts_2024,
  title={Scaling llm test-time compute optimally can be more effective than scaling model parameters},
  author={Snell, Charlie and Lee, Jaehoon and Xu, Kelvin and Kumar, Aviral},
  journal={arXiv preprint arXiv:2408.03314},
  year={2024}
}

@inproceedings{tts_2025,
  title={s1: Simple test-time scaling},
  author={Muennighoff, Niklas and Yang, Zitong and Shi, Weijia and Li, Xiang Lisa and Fei-Fei, Li and Hajishirzi, Hannaneh and Zettlemoyer, Luke and Liang, Percy and Cand{\`e}s, Emmanuel and Hashimoto, Tatsunori B},
  booktitle={Proceedings of the 2025 Conference on Empirical Methods in Natural Language Processing},
  pages={20286--20332},
  year={2025}
}

@article{sc,
  title={Self-consistency improves chain of thought reasoning in language models},
  author={Wang, Xuezhi and Wei, Jason and Schuurmans, Dale and Le, Quoc and Chi, Ed and Narang, Sharan and Chowdhery, Aakanksha and Zhou, Denny},
  journal={arXiv preprint arXiv:2203.11171},
  year={2022}
}

@inproceedings{ac,
  title={Let’s sample step by step: Adaptive-consistency for efficient reasoning and coding with llms},
  author={Aggarwal, Pranjal and Madaan, Aman and Yang, Yiming and others},
  booktitle={Proceedings of the 2023 Conference on Empirical Methods in Natural Language Processing},
  pages={12375--12396},
  year={2023}
}

@article{esc,
  title={Escape sky-high cost: Early-stopping self-consistency for multi-step reasoning},
  author={Li, Yiwei and Yuan, Peiwen and Feng, Shaoxiong and Pan, Boyuan and Wang, Xinglin and Sun, Bin and Wang, Heda and Li, Kan},
  journal={arXiv preprint arXiv:2401.10480},
  year={2024}
}

@inproceedings{dsc,
  title={Make every penny count: Difficulty-adaptive self-consistency for cost-efficient reasoning},
  author={Wang, Xinglin and Feng, Shaoxiong and Li, Yiwei and Yuan, Peiwen and Zhang, Yueqi and Tan, Chuyi and Pan, Boyuan and Hu, Yao and Li, Kan},
  booktitle={Findings of the Association for Computational Linguistics: NAACL 2025},
  pages={6904--6917},
  year={2025}
}

@article{diff_aware_cot,
  title={Less is More Tokens: Efficient Math Reasoning via Difficulty-Aware Chain-of-Thought Distillation},
  author={Waheed, Abdul and Mitra, Chancharik and Wang, Laurie Z and Ramanan, Deva and Raj, Bhiksha},
  journal={arXiv preprint arXiv:2509.05226},
  year={2025}
}

@inproceedings{llm_already_knows,
  title={The llm already knows: Estimating llm-perceived question difficulty via hidden representations},
  author={Zhu, Yubo and Liu, Dongrui and Lin, Zecheng and Tong, Wei and Zhong, Sheng and Shao, Jing},
  booktitle={Proceedings of the 2025 Conference on Empirical Methods in Natural Language Processing},
  pages={1160--1176},
  year={2025}
}

@article{diffadapt,
  title={DiffAdapt: Difficulty-Adaptive Reasoning for Token-Efficient LLM Inference},
  author={Liu, Xiang and Hu, Xuming and Chu, Xiaowen and Choi, Eunsol},
  journal={arXiv preprint arXiv:2510.19669},
  year={2025}
}

@article{probing_difficulty,
  title={Probing the Difficulty Perception Mechanism of Large Language Models},
  author={Lee, Sunbowen and Yin, Qingyu and Leong, Chak Tou and Zhang, Jialiang and Gong, Yicheng and Ni, Shiwen and Yang, Min and Shen, Xiaoyu},
  journal={arXiv preprint arXiv:2510.05969},
  year={2025}
}

@inproceedings{last_embed_literary,
  title={What’s in a prompt? Language models encode literary style in prompt embeddings},
  author={Sarfati, Rapha{\"e}l and Moller, Haley and Liu, Toni JB and Boull{\'e}, Nicolas and Earls, Christopher},
  booktitle={Proceedings of the 2025 Conference on Empirical Methods in Natural Language Processing},
  pages={24070--24079},
  year={2025}
}

@inproceedings{last_embed_instruct,
  title={Do LLMs``know''internally when they follow instructions?},
  author={Heo, Juyeon and Heinze-Deml, Christina and Elachqar, Oussama and Chan, Kwan Ho Ryan and Ren, Shirley and Miller, Andrew and Nallasamy, Udhyakumar and Narain, Jaya},
  booktitle={International Conference on Learning Representations},
  volume={2025},
  pages={81339--81357},
  year={2025}
}

@article{last_embed_safety,
  title={Enhancing Safety of Large Language Models via Embedding Space Separation},
  author={Zhao, Xu and Wang, Xiting and Shen, Weiran},
  journal={arXiv preprint arXiv:2603.20206},
  year={2026}
}

@article{shannon_entropy,
  title={A mathematical theory of communication},
  author={Shannon, Claude Elwood},
  journal={The Bell system technical journal},
  volume={27},
  number={3},
  pages={379--423},
  year={1948},
  publisher={Nokia Bell Labs}
}

@article{ppo,
  title={Proximal policy optimization algorithms},
  author={Schulman, John and Wolski, Filip and Dhariwal, Prafulla and Radford, Alec and Klimov, Oleg},
  journal={arXiv preprint arXiv:1707.06347},
  year={2017}
}

@article{grpo,
  title={Deepseekmath: Pushing the limits of mathematical reasoning in open language models},
  author={Shao, Zhihong and Wang, Peiyi and Zhu, Qihao and Xu, Runxin and Song, Junxiao and Bi, Xiao and Zhang, Haowei and Zhang, Mingchuan and Li, YK and Wu, Yang and others},
  journal={arXiv preprint arXiv:2402.03300},
  year={2024}
}

@inproceedings{adaptthink,
  title={Adaptthink: Reasoning models can learn when to think},
  author={Zhang, Jiajie and Lin, Nianyi and Hou, Lei and Feng, Ling and Li, Juanzi},
  booktitle={Proceedings of the 2025 Conference on Empirical Methods in Natural Language Processing},
  pages={3716--3730},
  year={2025}
}

@article{math,
  title={Measuring mathematical problem solving with the math dataset},
  author={Hendrycks, Dan and Burns, Collin and Kadavath, Saurav and Arora, Akul and Basart, Steven and Tang, Eric and Song, Dawn and Steinhardt, Jacob},
  journal={arXiv preprint arXiv:2103.03874},
  year={2021}
}

@inproceedings{math500,
  title={Let's verify step by step},
  author={Lightman, Hunter and Kosaraju, Vineet and Burda, Yuri and Edwards, Harrison and Baker, Bowen and Lee, Teddy and Leike, Jan and Schulman, John and Sutskever, Ilya and Cobbe, Karl},
  booktitle={International Conference on Learning Representations},
  volume={2024},
  pages={39578--39601},
  year={2024}
}

@misc{amc23,
  author = {{AI-MO}},
  title = {Aimo validation amc dataset on hugging face},
  year = {2024},
  url = {https://huggingface.co/datasets/AI-MO/aimo-validation-amc}
}

@article{gpqa,
  title={Gpqa: A graduate-level google-proof q\&a benchmark},
  author={Rein, David and Hou, Betty Li and Stickland, Asa Cooper and Petty, Jackson and Pang, Richard Yuanzhe and Dirani, Julien and Michael, Julian and Bowman, Samuel R},
  journal={arXiv preprint arXiv:2311.12022},
  year={2023}
}

@misc{aime24,
      title={American Invitational Mathematics Examination (AIME) 2024}, 
      author={Zhang, Yifan and Math-AI, Team},
      year={2024},
}

@misc{aime25,
      title={American Invitational Mathematics Examination (AIME) 2025}, 
      author={Zhang, Yifan and Math-AI, Team},
      year={2025},
}

@article{mmlu_pro,
  title={Mmlu-pro: A more robust and challenging multi-task language understanding benchmark},
  author={Wang, Yubo and Ma, Xueguang and Zhang, Ge and Ni, Yuansheng and Chandra, Abhranil and Guo, Shiguang and Ren, Weiming and Arulraj, Aaran and He, Xuan and Jiang, Ziyan and others},
  journal={Advances in Neural Information Processing Systems},
  volume={37},
  pages={95266--95290},
  year={2024}
}

@article{qwen2.5,
  title={Qwen2.5 Technical Report},
  author={Qwen An Yang and Baosong Yang and Beichen Zhang and Binyuan Hui and Bo Zheng and Bowen Yu and Chengyuan Li and Dayiheng Liu and Fei Huang and Guanting Dong and Haoran Wei and Huan Lin and Jian Yang and Jianhong Tu and Jianwei Zhang and Jianxin Yang and Jiaxin Yang and Jingren Zhou and Junyang Lin and Kai Dang and Keming Lu and Keqin Bao and Kexin Yang and Le Yu and Mei Li and Mingfeng Xue and Pei Zhang and Qin Zhu and Rui Men and Runji Lin and Tianhao Li and Tingyu Xia and Xingzhang Ren and Xuancheng Ren and Yang Fan and Yang Su and Yi-Chao Zhang and Yunyang Wan and Yuqi Liu and Zeyu Cui and Zhenru Zhang and Zihan Qiu and Shanghaoran Quan and Zekun Wang},
  journal={ArXiv},
  year={2024},
  volume={abs/2412.15115},
  url={https://api.semanticscholar.org/CorpusID:274859421}
}

@article{gemma_3,
  title={Gemma 3 Technical Report},
  author={Gemma Team Aishwarya Kamath and Johan Ferret and Shreya Pathak and Nino Vieillard and Ramona Merhej and Sarah Perrin and Tatiana Matejovicova and Alexandre Ram'e and Morgane Rivi{\`e}re and Louis Rouillard and Thomas Mesnard and Geoffrey Cideron and Jean-Bastien Grill and Sabela Ramos and Edouard Yvinec and Michelle Casbon and Etienne Pot and Ivo Penchev and Gael Liu and Francesco Visin and Kathleen Kenealy and Lucas Beyer and Xiaohai Zhai and Anton Tsitsulin and R{\'o}bert Istvan Busa-Fekete and Alex Feng and Noveen Sachdeva and Benjamin Coleman and Yi Gao and Basil Mustafa and Iain Barr and Emilio Parisotto and David Tian and Matan Eyal and Colin Cherry and Jan-Thorsten Peter and Danila Sinopalnikov and Surya Bhupatiraju and Rishabh Agarwal and Mehran Kazemi and Dan Malkin and Ravin Kumar and David Vilar and Idan Brusilovsky and Jiaming Luo and Andreas Steiner and Abe Friesen and Abhanshu Sharma and Abheesht Sharma and Adi Mayrav Gilady and Adrian Goedeckemeyer and Alaa Saade and Alexander Kolesnikov and Alexei Bendebury and Alvin Abdagic and Amit Vadi and Andr'as Gyorgy and Andr{\'e} Susano Pinto and Anil Das and Ankur Bapna and Antoine Miech and Antoine Yang and Antonia Paterson and Ashish Shenoy and Ayan Chakrabarti and Bilal Piot and Boxi Wu and Bobak Shahriari and Bryce Petrini and Charlie Chen and Charline Le Lan and Christopher A. Choquette-Choo and Cj Carey and Cormac Brick and Daniel Deutsch and Danielle Eisenbud and Dee Cattle and Derek Zhiyuan Cheng and Dimitris Paparas and Divyashree Shivakumar Sreepathihalli and Doug Reid and Dustin Tran and Dustin Zelle and Eric Noland and Erwin Huizenga and Eugene Kharitonov and Frederick Liu and Gagik Amirkhanyan and Glenn Cameron and Hadi Hashemi and Hanna Klimczak-Pluci'nska and Harman Singh and Harsh Mehta and Harshal Tushar Lehri and Hussein Hazimeh and Ian Ballantyne and Idan Szpektor and Ivan Nardini and Jean Pouget-Abadie and Jetha Chan and Joe Stanton and J. Michael Wieting and Jonathan Lai and Jordi Orbay and Joe Fernandez and Joshua Newlan and Junsong Ji and Jyotinder Singh and Kat Black and Kathy Yu and Kevin Hui and Kiran Vodrahalli and Klaus Greff and Linhai Qiu and Marcella Valentine and Marina Coelho and Marvin Ritter and Matt Hoffman and Matthew Watson and Mayank Chaturvedi and Michael Moynihan and Min Ma and Nabila Babar and Natasha Noy and Nathan Byrd and Nick Roy and Nikola Momchev and Nilay Chauhan and Oskar Bunyan and Pankil Botarda and Paul Caron and Paul Kishan Rubenstein and Phil Culliton and Philipp Schmid and Pier Giuseppe Sessa and Ping-mei Xu and Piotr Stańczyk and Pouya Dehghani Tafti and Rakesh Shivanna and Renjie Wu and Renke Pan and Reza Ardeshir Rokni and Rob Willoughby and Rohith Vallu and Ryan Mullins and Sammy Jerome and Sara Smoot and Sertan Girgin and Shariq Iqbal and Shashir Reddy and Shruti Sheth and Siim P{\~o}der and Sijal Bhatnagar and Sindhu Raghuram Panyam and Sivan Eiger and Susan Zhang and Tianqi Liu and Trevor Yacovone and Tyler Liechty and Uday Kalra and Utku Evci and Vedant Misra and Vincent Roseberry and Vladimir Feinberg and Vlad Kolesnikov and Woohyun Han and Woosuk Kwon and Xi Chen and Yinlam Chow and Yuvein Zhu and Zichuan Wei and Zoltan Egyed and Victor Cotruta and Minh Giang and Phoebe Kirk and Anand Rao and Jessica Lo and Erica Moreira and Luiz Gustavo Martins and Omar Sanseviero and Lucas Gonzalez and Zach Gleicher and Tris Warkentin and Vahab S. Mirrokni and Evan Senter and Eli Collins and Joelle Barral and Zoubin Ghahramani and Raia Hadsell and Yossi Matias and D. Sculley and Slav Petrov and Noah Fiedel and Noam Shazeer and Oriol Vinyals and Jeffrey Dean and Demis Hassabis and Koray Kavukcuoglu and Cl{\'e}ment Farabet and Elena Buchatskaya and Jean-Baptiste Alayrac and Rohan Anil and Dmitry Lepikhin and Sebastian Borgeaud and Olivier Bachem and Armand Joulin and Alek Andreev and Cassidy Hardin and Robert Dadashi and L'eonard Hussenot},
  journal={ArXiv},
  year={2025},
  volume={abs/2503.19786},
  url={https://api.semanticscholar.org/CorpusID:277313563}
}

@article{analyis_prob_diff1,
  title={Towards thinking-optimal scaling of test-time compute for llm reasoning},
  author={Yang, Wenkai and Ma, Shuming and Lin, Yankai and Wei, Furu},
  journal={Advances in Neural Information Processing Systems},
  volume={38},
  pages={43605--43631},
  year={2026}
}

@inproceedings{analysis_prob_diff2,
  title={Step-kto: Optimizing mathematical reasoning through stepwise binary feedback},
  author={Lin, Yen-Ting and Jin, Di and Xu, Tengyu and Wu, Tianhao and Sukhbaatar, Sainbayar and Zhu, Chen and He, Yun and Chen, Yun-Nung and Weston, Jason E and Tian, Yuandong and others},
  booktitle={Proceedings of The 3rd Workshop on Mathematical Natural Language Processing (MathNLP 2025)},
  pages={15--33},
  year={2025}
}

\appendix
\label{sec:appendix}
\section{Full Prompts for Reasoning Chain Generation}
\label{sec:appendix Full Prompt}
\begin{figure}[ht]
  \includegraphics[width=\columnwidth]{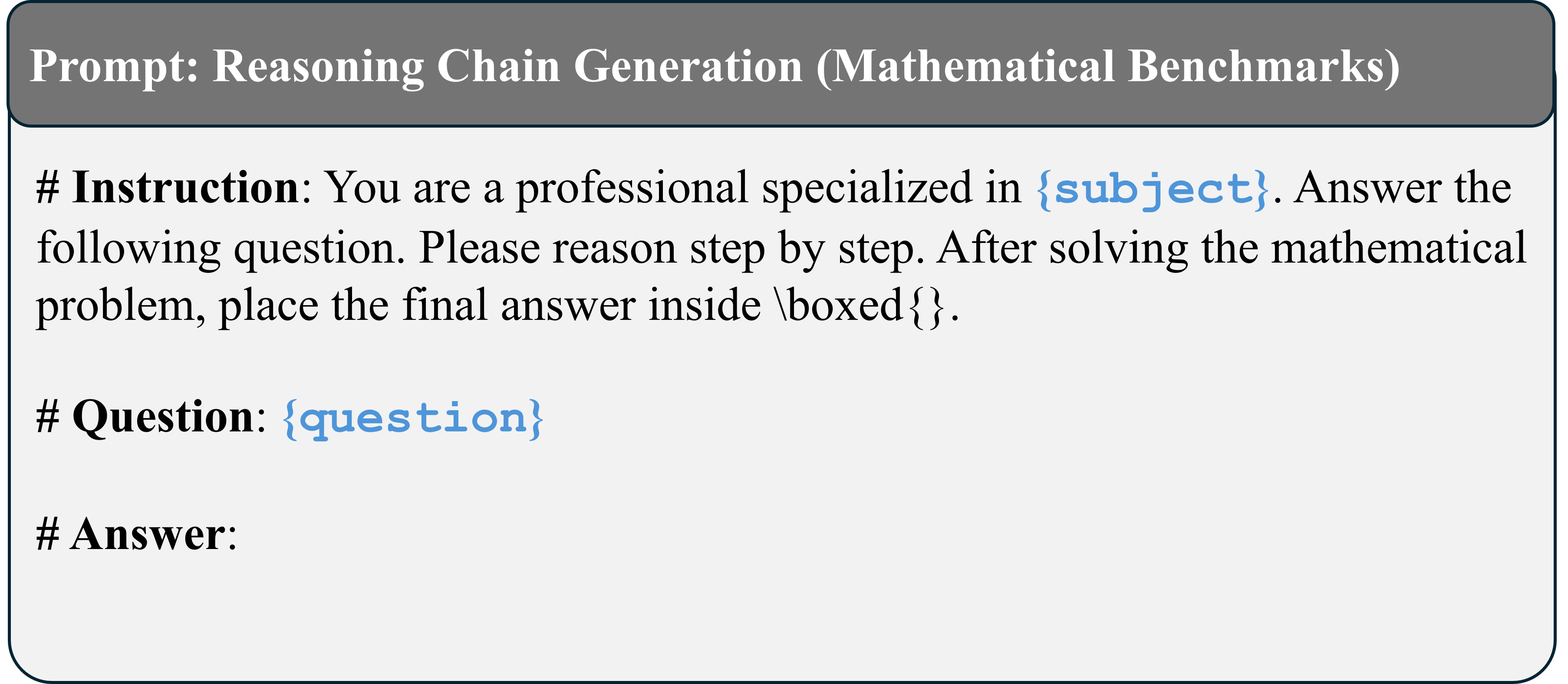}
  \caption{Prompt used for mathematical reasoning problems.}
  \label{fig: prompt_math}
\end{figure}

\begin{figure}[ht]
  \includegraphics[width=\columnwidth]{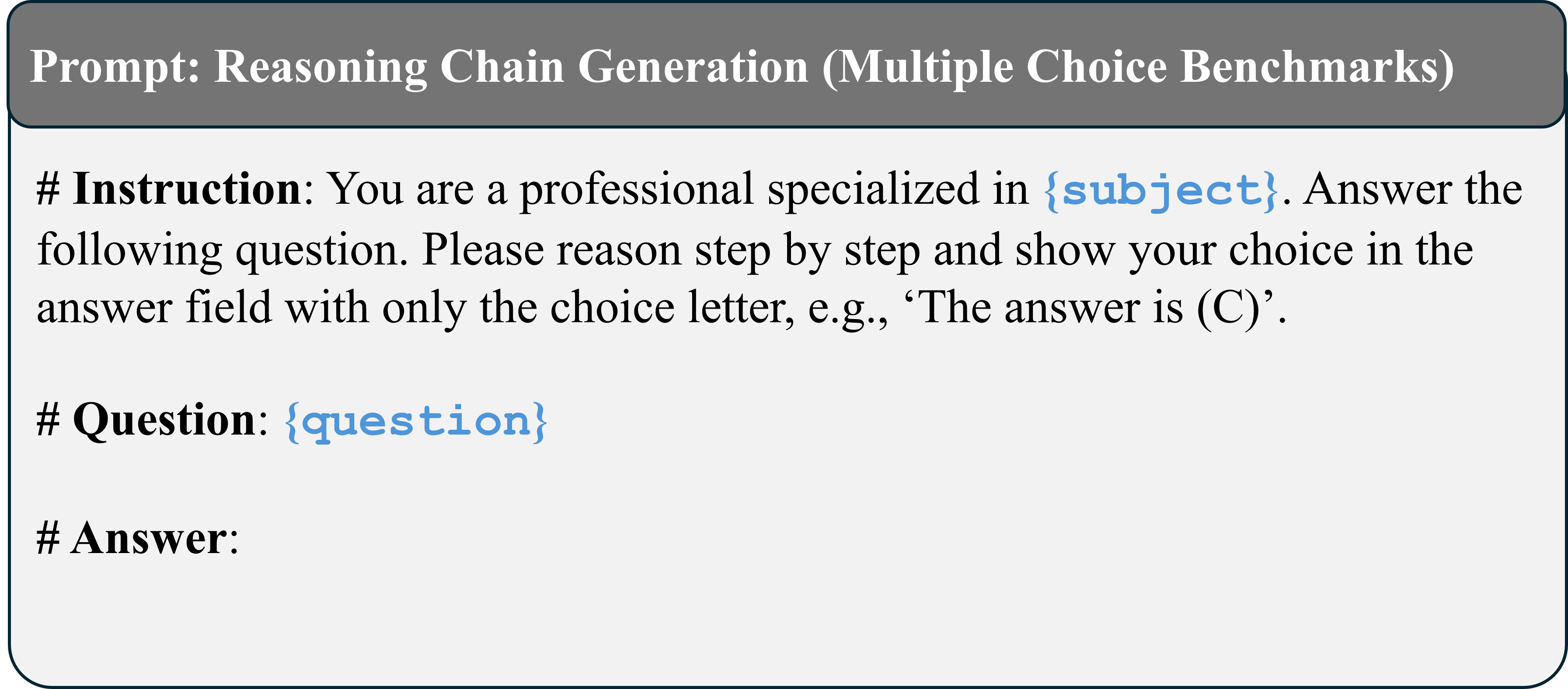}
  \caption{Prompt used for multiple-choice problems.}
  \label{fig: prompt_mcq}
\end{figure}

This section presents the full prompts used to generate reasoning chains. All experiments are conducted using the same prompts, enabling a fair comparison between SC and FSC based on the generated reasoning paths. Figure~\ref{fig: prompt_math} shows the prompt used for open-ended problems (MATH500, AMC23, AIME 2024, and AIME 2025), while Figure~\ref{fig: prompt_mcq} shows the prompt used for multiple-choice problems (GPQA-Diamond and MMLU-Pro). Under this setup, all methods generate reasoning chains under the same input conditions, allowing us to compare the effect of the inference strategy itself.

\begin{figure*}[t]
  \centering
  \begin{subfigure}[t]{0.32\textwidth}
    \centering
    \includegraphics[width=\linewidth]{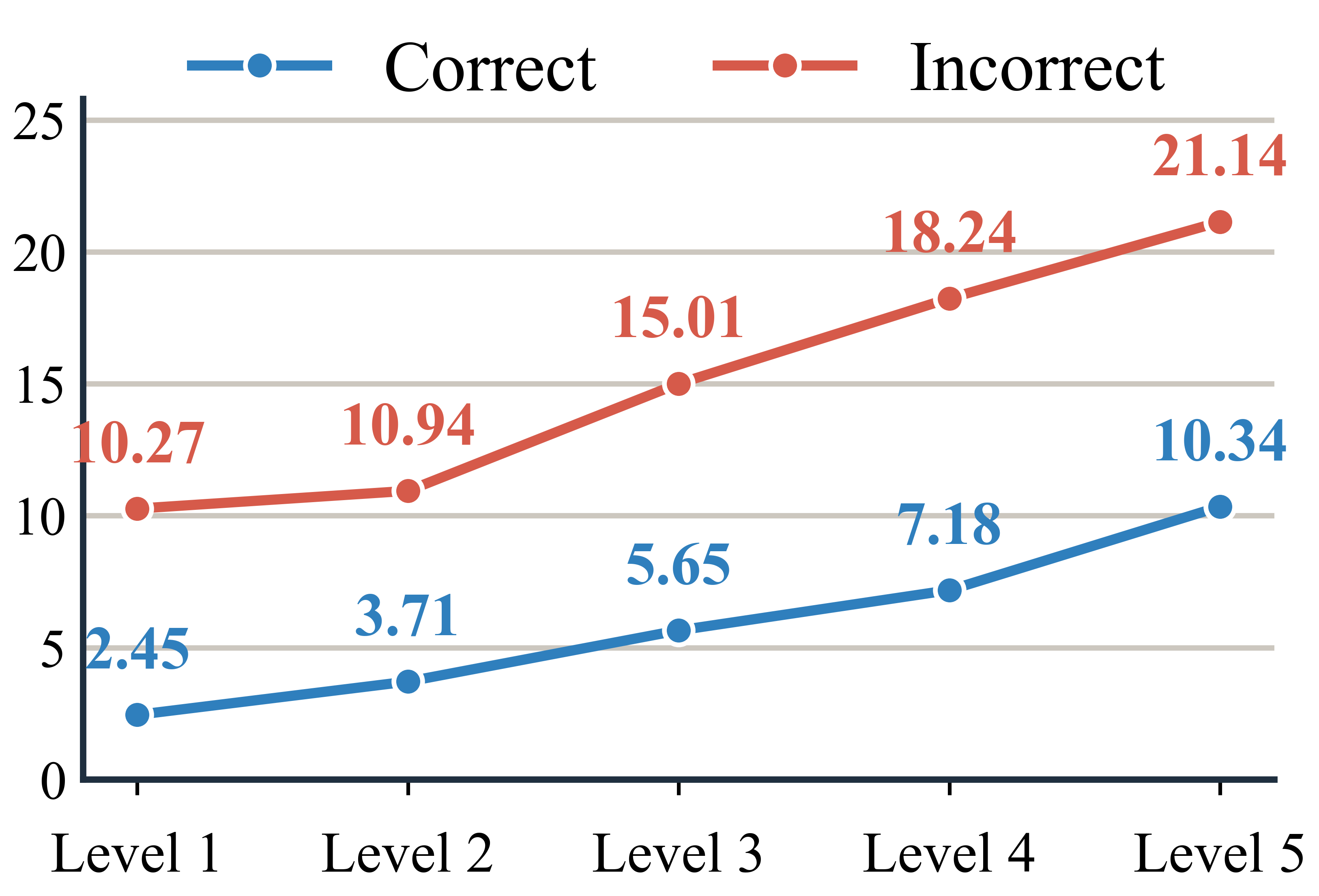}
    \caption{Qwen2.5-3B}
    \label{fig: answer_diversity_math_by_level_qwen2.5-3b-instruct}
  \end{subfigure}\hfill
  \begin{subfigure}[t]{0.32\textwidth}
    \centering
    \includegraphics[width=\linewidth]{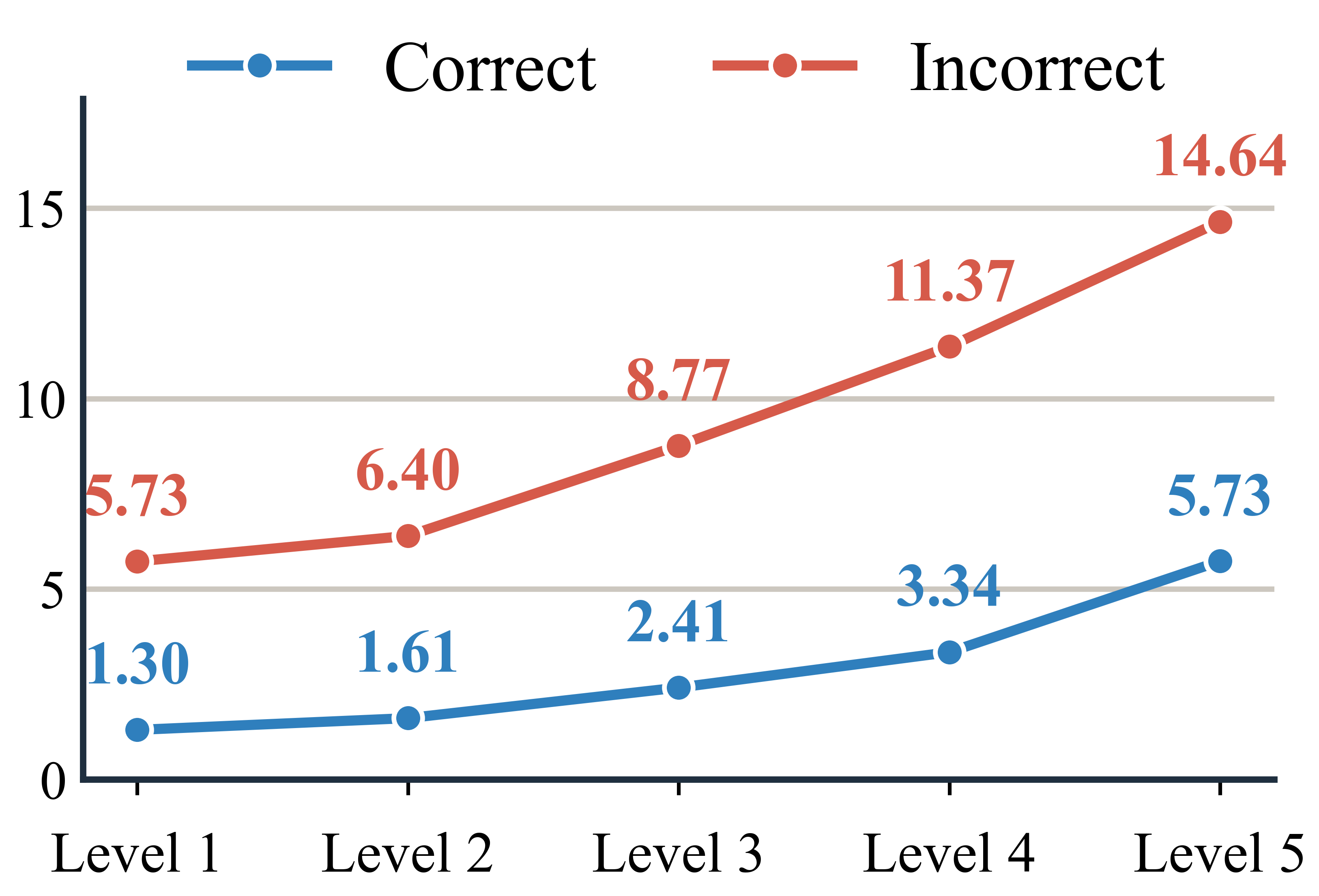}
    \caption{Qwen2.5-14B}
    \label{fig: answer_diversity_math_by_level_qwen2.5-14b-instruct}
  \end{subfigure}\hfill
  \begin{subfigure}[t]{0.32\textwidth}
    \centering
    \includegraphics[width=\linewidth]{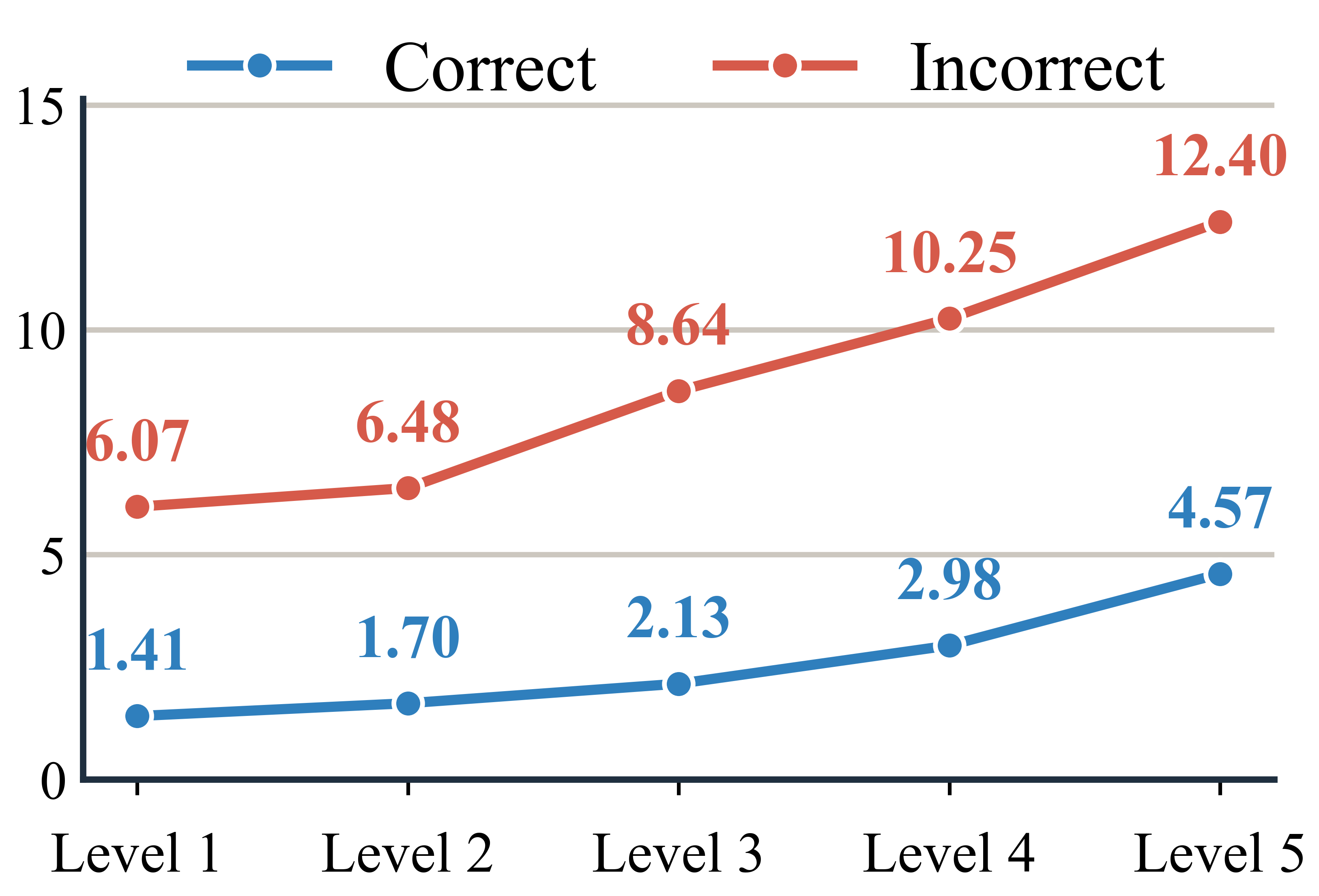}
    \caption{Gemma-3-4B}
    \label{fig: answer_diversity_math_by_level_gemma}
  \end{subfigure}
  \caption{\textbf{Answer diversity across models.} All models show increasing answer diversity as problem difficulty increases, particularly for incorrect predictions, supporting the use of output entropy as a continuous difficulty signal.}
  \label{fig: answer_diversity_math_by_level_all}
\end{figure*}

\begin{table*}[t]
\centering
\small
\renewcommand{\arraystretch}{1.0}
\resizebox{\linewidth}{!}{%
\begin{tabular}{lllrl}
\toprule
Dataset & Domain & Answer Format & \# Samples & License \\
\midrule

\multicolumn{5}{c}{\cellcolor{Oursgray}\textit{Mathematical Reasoning}} \\
\midrule
MATH Train   & Math reasoning   & Arabic number & 7,500 & MIT License \\
MATH500      & Math reasoning   & Arabic number & 500   & MIT License \\
AMC23        & Math competition & Arabic number & 50    & Apache License 2.0 \\
AIME2024     & Math competition & Arabic number & 30    & CC BY-NC-SA 4.0 \\
AIME2025     & Math competition & Arabic number & 30    & CC BY-NC-SA 4.0 \\

\midrule
\multicolumn{5}{c}{\cellcolor{Oursgray}\textit{Scientific Reasoning}} \\
\midrule
GPQA-Diamond & STEM QA          & Option (A--D) & 198   & MIT License \\

\midrule
\multicolumn{5}{c}{\cellcolor{Oursgray}\textit{General-Domain Reasoning}} \\
\midrule
MMLU-Pro     & Multi-domain QA  & Option (A--J) & 12,032 & MIT License \\

\bottomrule
\end{tabular}%
}
\caption{
Dataset statistics and license information. 
The \# Samples column denotes the number of questions used in our experiments.
}
\label{tab:dataset_information}
\end{table*}

\section{Motivation Analysis on Entropy-Difficulty Relationship}
\label{sec:appendix motivation}
We further examine whether the relationship between entropy and problem difficulty observed in \S\ref{sec: Motivation} is not limited to a specific model, but appears consistently across diverse models. To this end, in addition to Qwen2.5-7B used in the main text, we conduct the same experiment on Qwen2.5-3B, Qwen2.5-14B, and Gemma-3-4B. For each model, we generate 40 reasoning chains for 7,500 problems from the MATH training set and measure the number of unique answers according to the difficulty level of each problem. We further analyze the results separately for correct and incorrect cases.

Figure~\ref{fig: answer_diversity_math_by_level_all} shows the results for Qwen2.5-3B, Qwen2.5-14B, and Gemma-3-4B. Across all models, we consistently observe that (1) incorrect cases exhibit higher answer diversity than correct cases, and (2) the number of unique answers gradually increases as the difficulty level of the problem increases. These results suggest that, regardless of model size or architecture, the diversity of the output distribution is closely associated with problem difficulty.

\section{Detailed Experimental Setup}
\label{sec:appendix detailed experimental setup}

\paragraph{Datasets.}
In this study, we use several benchmarks to evaluate mathematical reasoning, STEM-based question answering, and general-domain reasoning abilities across diverse academic fields. Table~\ref{tab:dataset_information} summarizes the domain, answer format, number of evaluation samples, and license information for each dataset used in this work.

\paragraph{Implementation Details.}
To construct the dataset for training the probe, we generate 40 reasoning paths for each of the 7,500 samples in the MATH training split. To obtain diverse reasoning paths, we set both temperature and top-p to 1.0.
During inference, we use temperature 0.7 and top-p 0.95 for more stable generation, and conduct all experiments with a single run. We set the stopping threshold of AC to 0.95 and the window size of ESC to 5. For DSC, we follow the original settings, except that we adjust the judge window size to 24.

\begin{figure*}[t]
  \centering
  \begin{subfigure}[t]{\linewidth}
    \centering
    \includegraphics[width=0.79\linewidth]{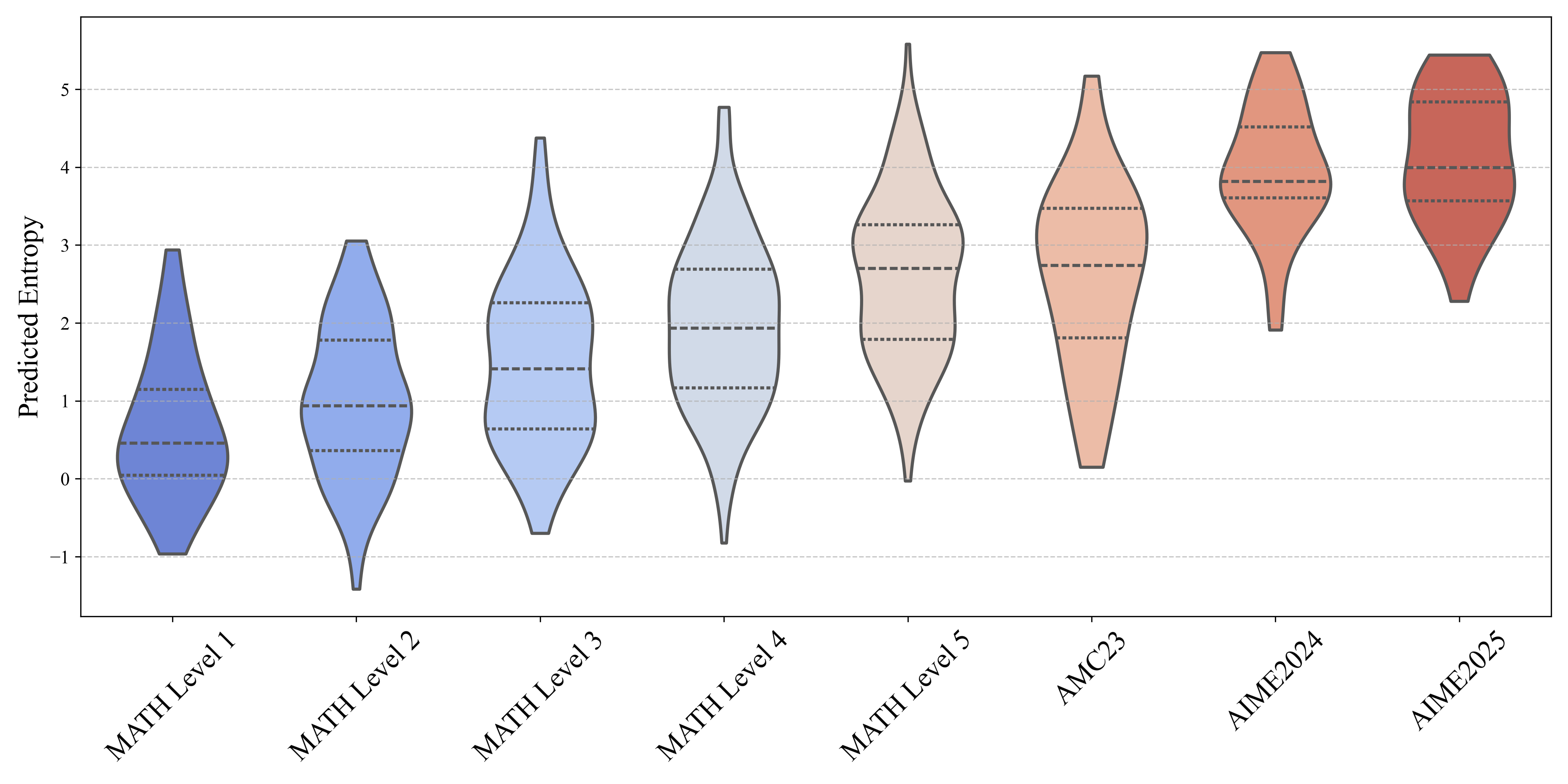}
    \caption{Qwen2.5-3B}
    \label{fig: entropy_violin_qwen2.5_3b}
  \end{subfigure}

  \vspace{0.9em}

  \begin{subfigure}[t]{\linewidth}
    \centering
    \includegraphics[width=0.79\linewidth]{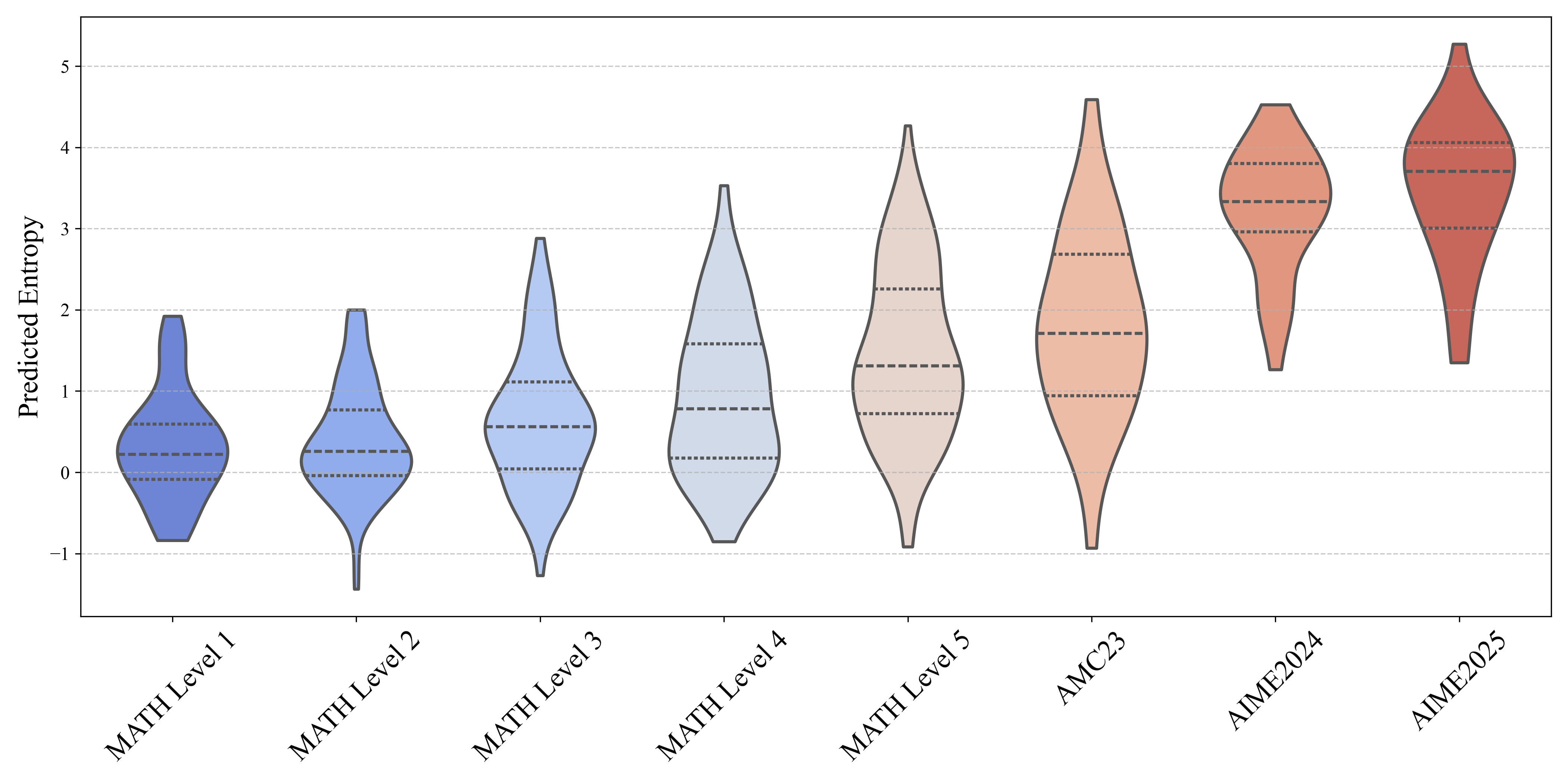}
    \caption{Qwen2.5-14B}
    \label{fig: entropy_violin_qwen2.5_14b}
  \end{subfigure}

  \vspace{0.9em}

  \begin{subfigure}[t]{\linewidth}
    \centering
    \includegraphics[width=0.79\linewidth]{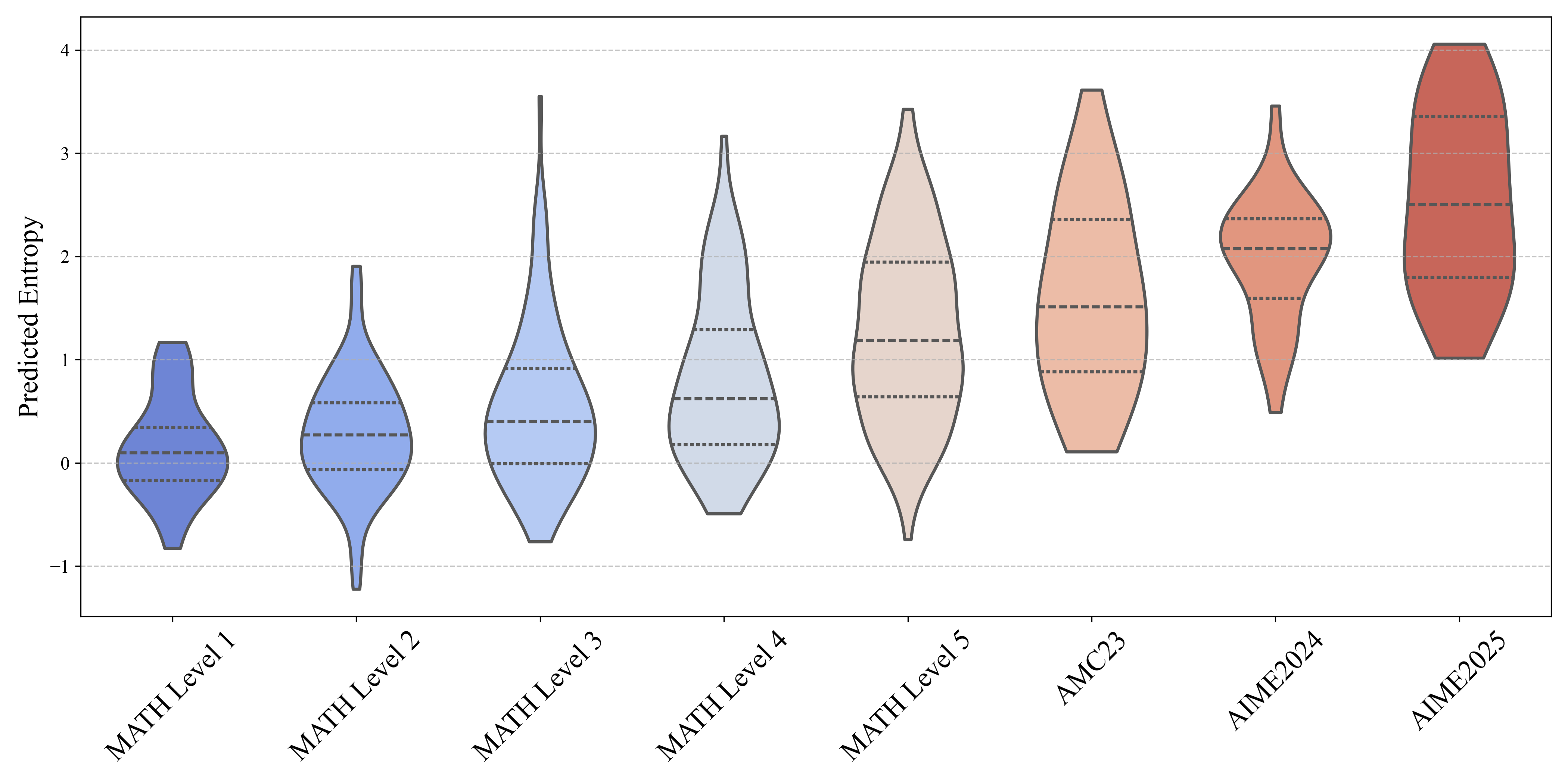}
    \caption{Gemma-3-4B}
    \label{fig: entropy_violin_gemma3_4b}
  \end{subfigure}
  \caption{Entropy distributions predicted by the probe across difficulty levels for various models.}
  \label{fig: entropy_violin_all}
\end{figure*}

\begin{figure*}[t]
  \centering
  \begin{subfigure}[t]{\linewidth}
    \centering
    \includegraphics[width=\linewidth]{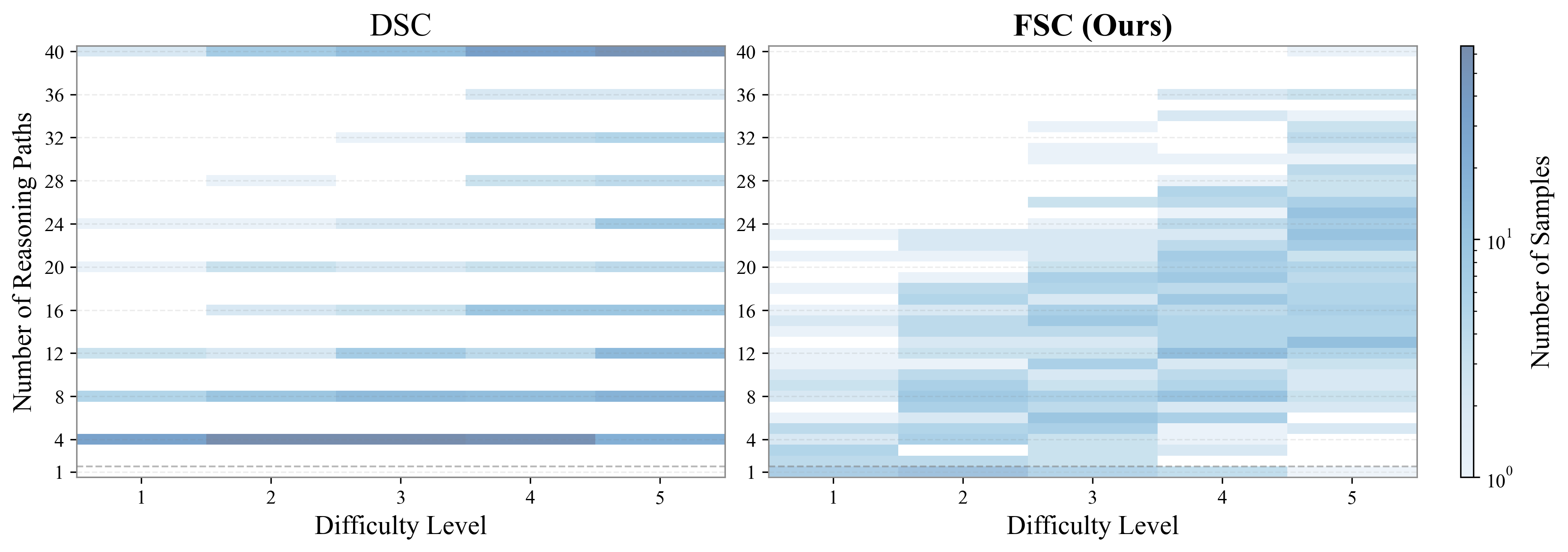}
    \caption{Qwen2.5-3B}
    \label{fig: path_allocation_qwen2.5_3b}
  \end{subfigure}

  \vspace{1.3em}

  \begin{subfigure}[t]{\linewidth}
    \centering
    \includegraphics[width=\linewidth]{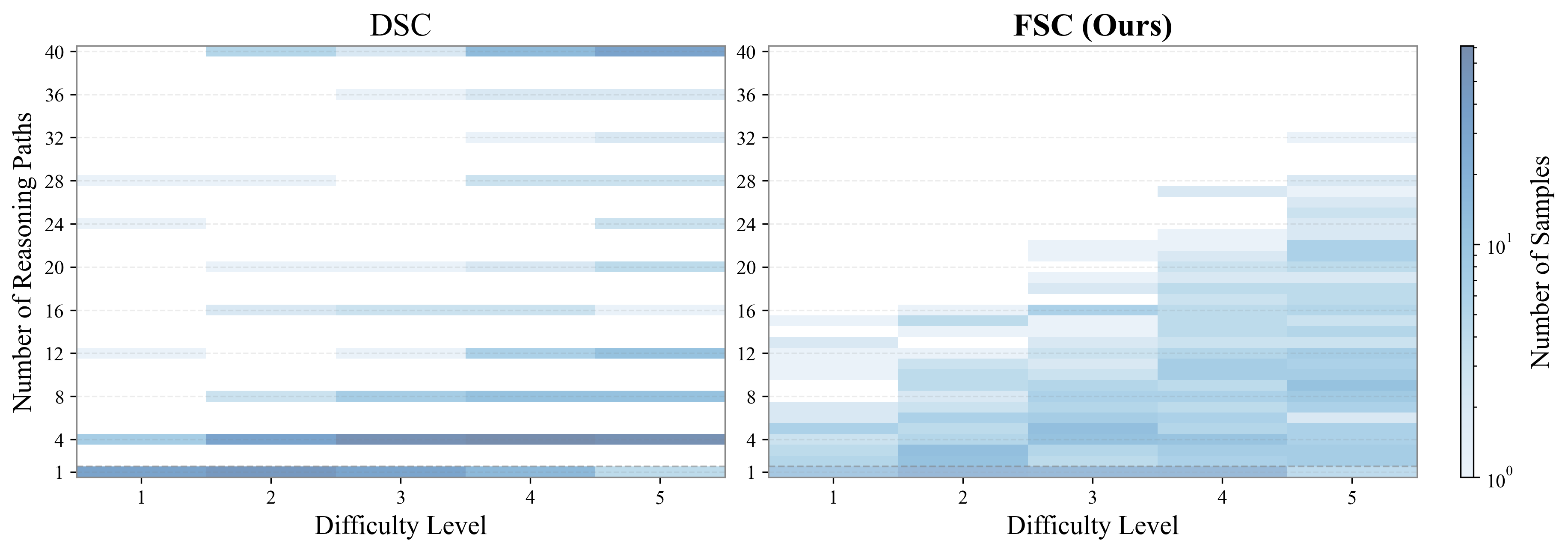}
    \caption{Qwen2.5-14B}
    \label{fig: path_allocation_qwen2.5_14b}
  \end{subfigure}

  \vspace{1.3em}

  \begin{subfigure}[t]{\linewidth}
    \centering
    \includegraphics[width=\linewidth]{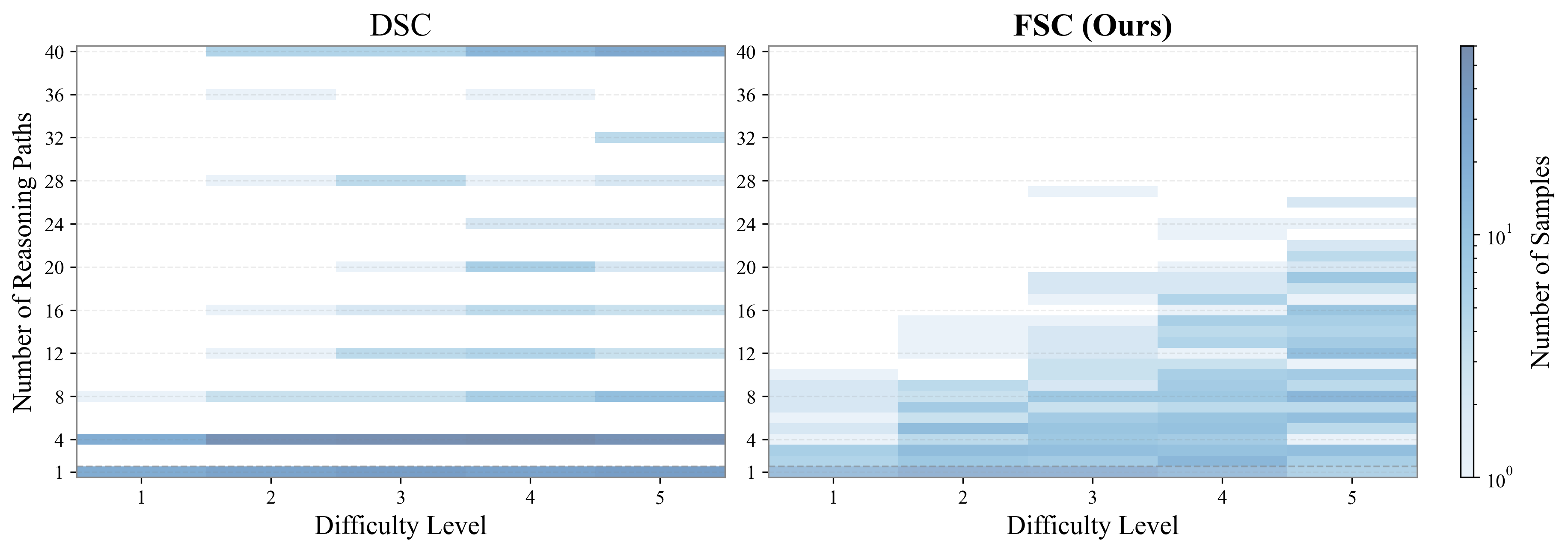}
    \caption{Gemma-3-4B}
    \label{fig: path_allocation_gemma3_4b}
  \end{subfigure}
  \caption{Distribution of the number of reasoning paths across difficulty levels for various models.}
  \label{fig: path_allocation_all}
\end{figure*}

\begin{figure*}[t]
  \centering
  \begin{subfigure}[t]{\linewidth}
    \centering
    \includegraphics[width=0.79\linewidth]{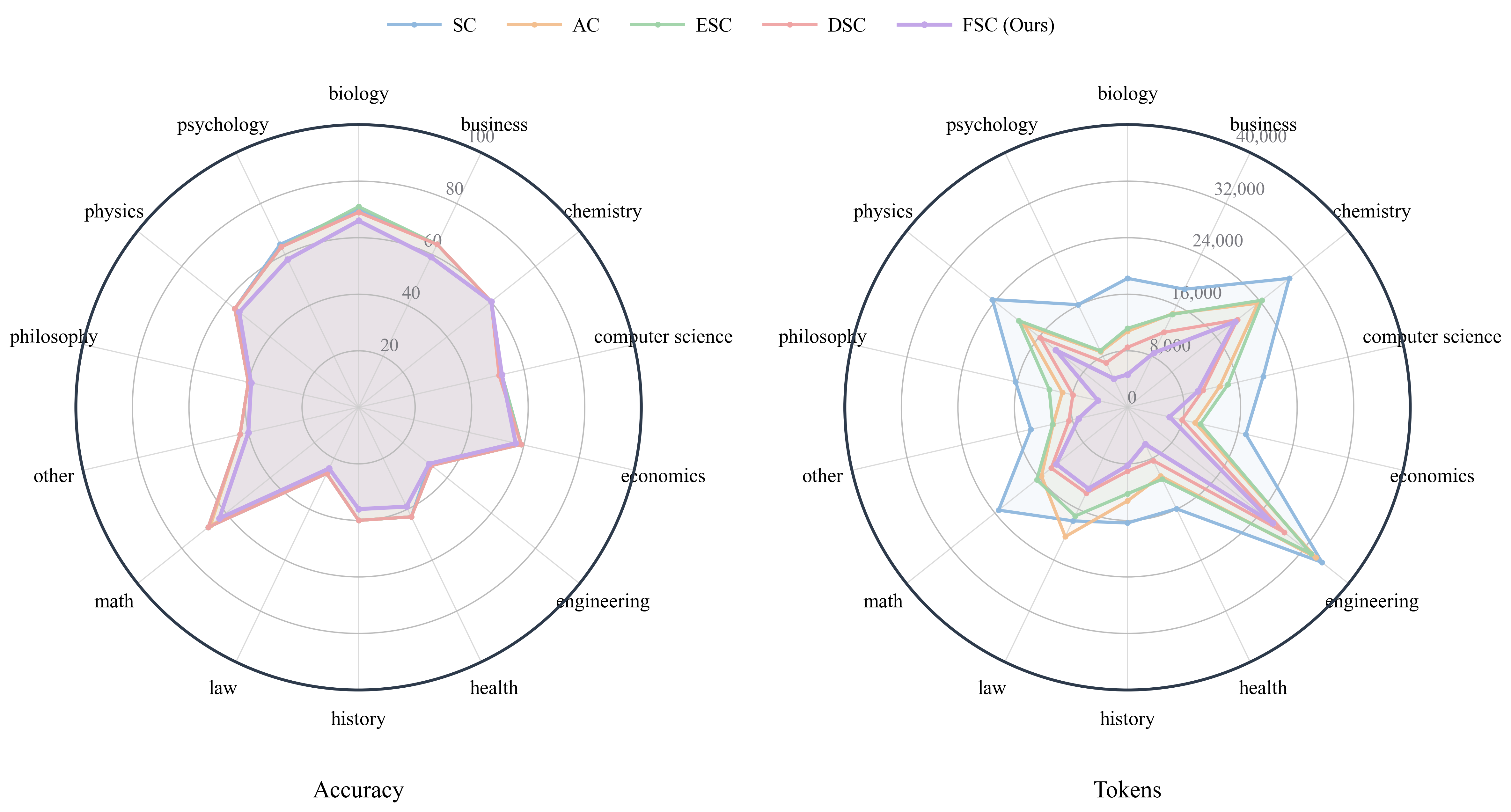}
    \caption{Qwen2.5-3B}
    \label{fig: mmlu_pro_category_radar_qwen2.5-3b-instruct}
  \end{subfigure}

  \vspace{0.9em}

  \begin{subfigure}[t]{\linewidth}
    \centering
    \includegraphics[width=0.79\linewidth]{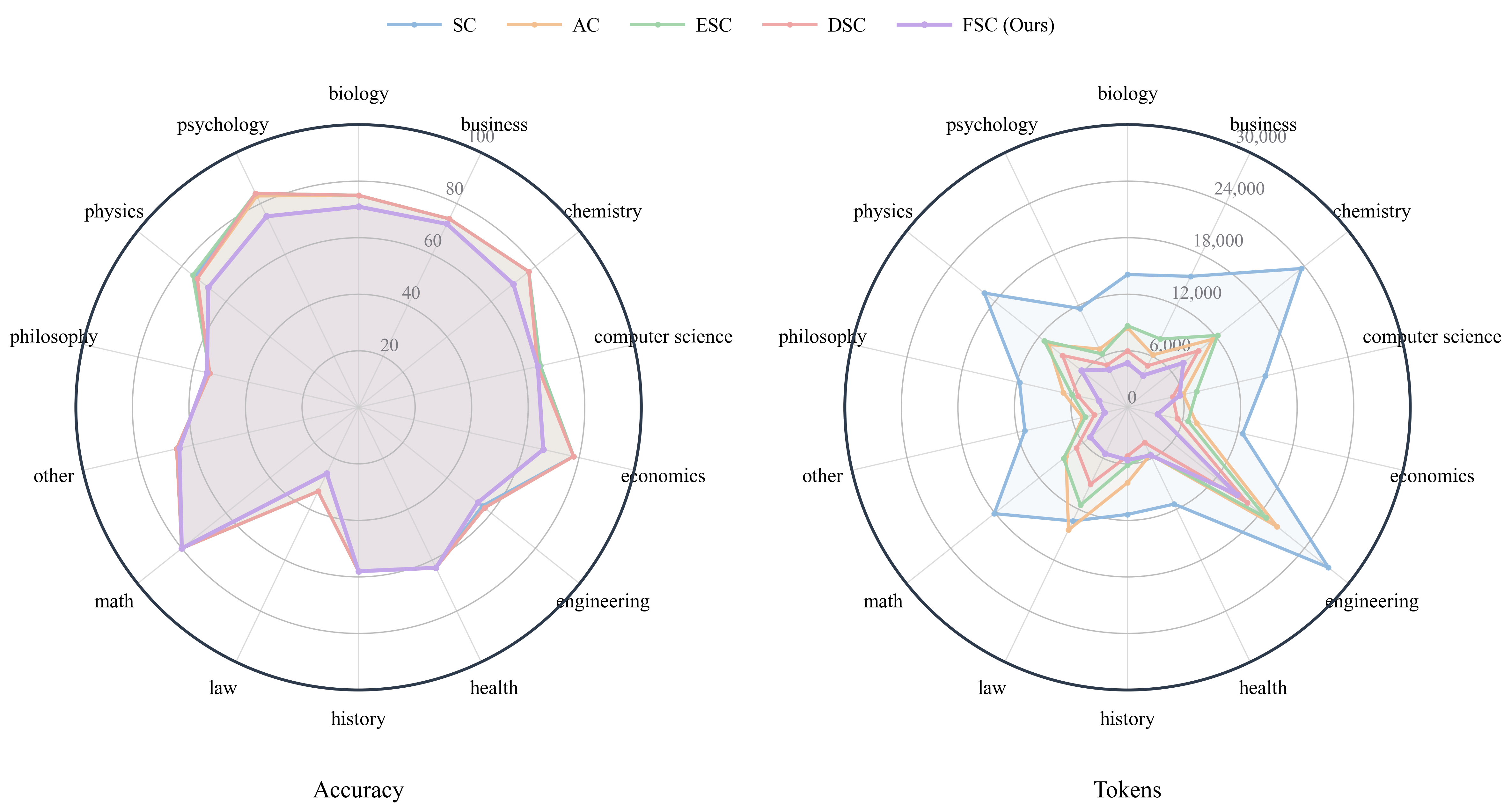}
    \caption{Qwen2.5-14B}
    \label{fig: mmlu_pro_category_radar_qwen2.5-14b-instruct}
  \end{subfigure}

  \vspace{0.9em}

  \begin{subfigure}[t]{\linewidth}
    \centering
    \includegraphics[width=0.79\linewidth]{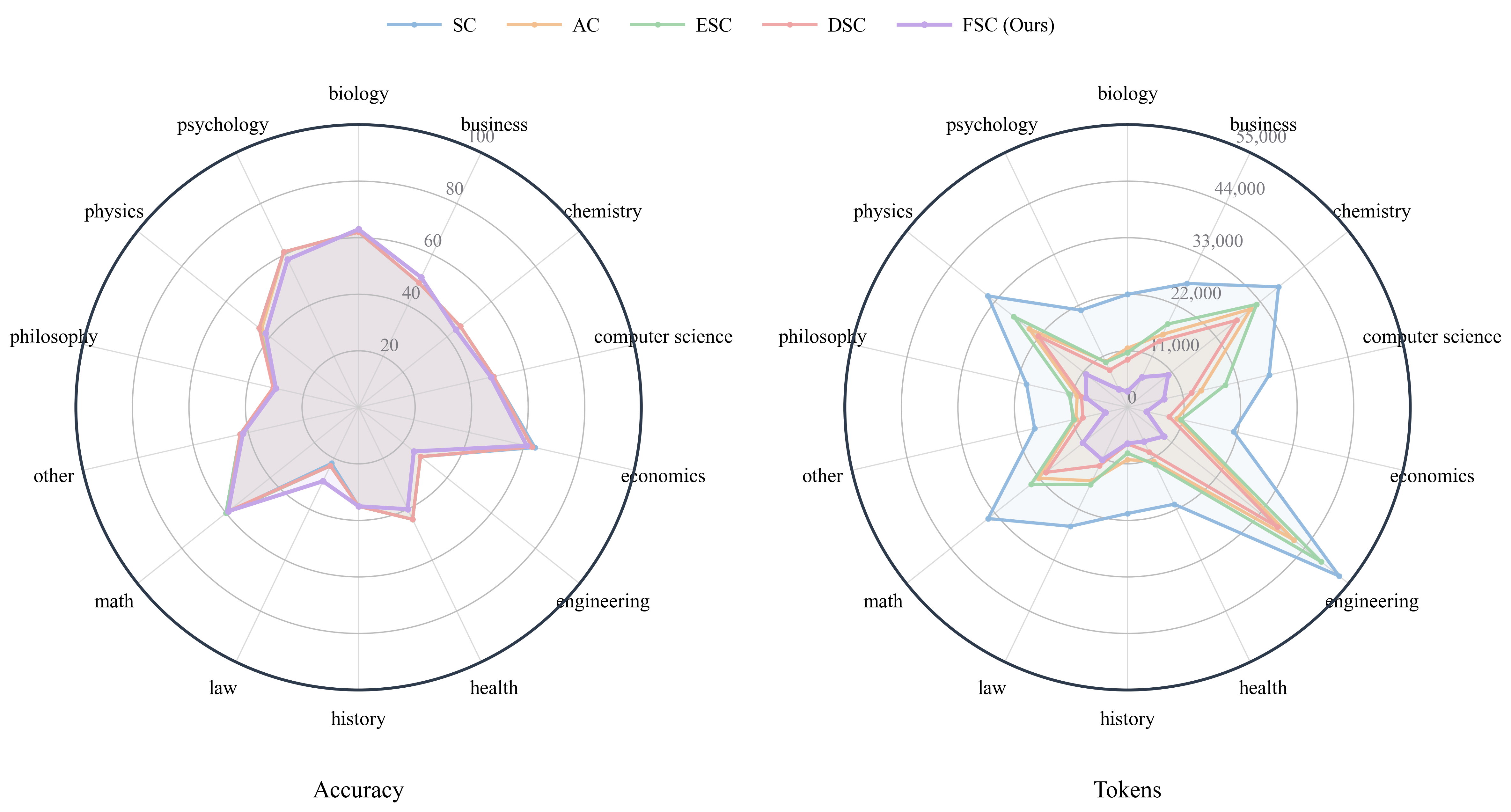}
    \caption{Gemma-3-4B}
    \label{fig: mmlu_pro_category_radar_gemma-3-4b-instruct}
  \end{subfigure}
  \caption{\textbf{Token efficiency comparison on MMLU-Pro across different models.} Panels compare FSC and baseline methods on Qwen2.5-3B, Qwen2.5-14B, and Gemma-3-4B, showing that FSC maintains strong token efficiency across model scales and architectures.}
  \label{fig: mmlu_pro_category_radar_all}
\end{figure*}

\begin{figure*}[t]
  \centering

  \begin{subfigure}[t]{\linewidth}
    \centering
    \includegraphics[width=\linewidth]{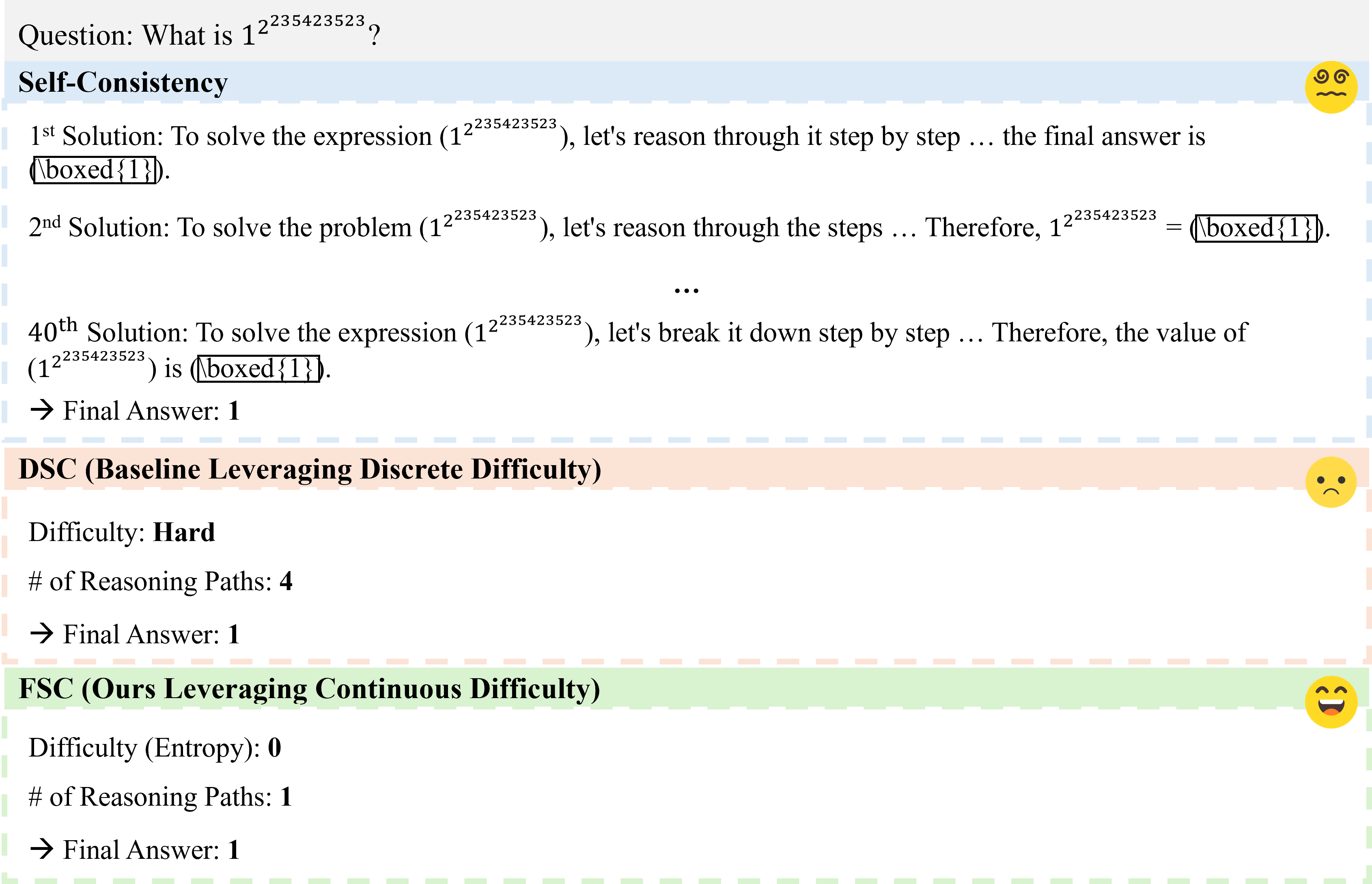}
    \caption{Easy problem (MATH500, Level 1)}
    \label{fig: case_study_easy}
  \end{subfigure}

  \vspace{1.5em}

  \begin{subfigure}[t]{\linewidth}
    \centering
    \includegraphics[width=\linewidth]{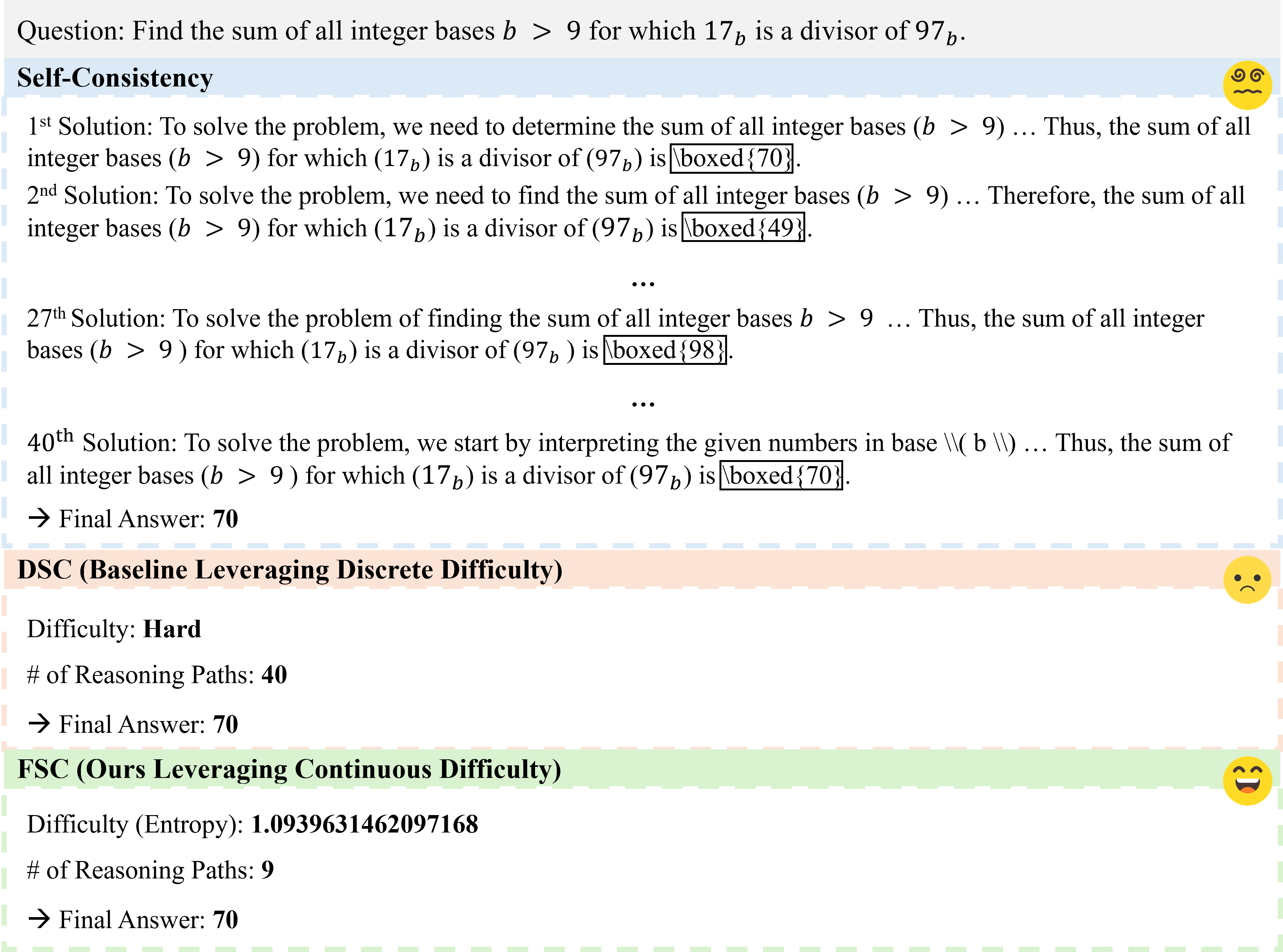}
    \caption{Challenging problem (AIME2025)}
    \label{fig: case_study_hard}
  \end{subfigure}

  \caption{Comparison of DSC and FSC on mathematical problems.}
  \label{fig: case_study}
\end{figure*}

\section{Case Study}
\label{sec: case_study}
Figure~\ref{fig: case_study} illustrates the difference between DSC and FSC in dynamically allocating inference paths. Figure~\ref{fig: case_study_easy} shows that when a relatively easy problem, such as one from MATH500 (Level 1), is given as input, all inference paths in SC converge to the same answer. Although this indicates that the problem is sufficiently easy from the model’s perspective, DSC incorrectly judges it as difficult and unnecessarily allocates four inference paths. In contrast, FSC accurately predicts the difficulty of the input problem using a probe, thereby substantially improving token efficiency while using only one inference path.

This trend is also observed for difficult problems, as shown in Figure~\ref{fig: case_study_hard}. Even for challenging problems such as those from AIME2025, FSC continuously estimates the difficulty perceived by the model and allocates fewer inference paths than DSC, thereby maximizing efficiency.

\section{Use of AI Tools}
\label{sec:appendix use of ai tools}
AI tools were used only for limited supportive purposes during the preparation of this manuscript. Specifically, AI tools such as OpenAI's ChatGPT were used to assist with translation, improve the clarity and fluency of the writing, and explore related keywords and expressions. However, the core ideas, methodological design, experimental implementation, result analysis, and final interpretations of this study were all independently conducted by the authors. In addition, all cited references included in this paper were directly verified by the authors, and no references were added solely based on content generated by AI tools.

\end{document}